\documentclass{article}

\usepackage{microtype}
\usepackage{paralist,enumitem}
\usepackage[table]{xcolor}

\usepackage{graphicx}
\usepackage{subfigure}
\usepackage{booktabs} % for professional tables
\usepackage{amsmath}
\usepackage{amsfonts}
\usepackage{nccmath}
\newtheorem{lemma}{Lemma}
\usepackage{textcomp}
\usepackage{upgreek}
\usepackage{diagbox}

\usepackage{bbm}
\usepackage{mathtools}
\usepackage{multirow} 

\DeclarePairedDelimiter\floor{\lfloor}{\rfloor}

\newtheorem{theorem}{Theorem}
\newtheorem{definition}{Definition}

\newtheorem{observation}{Observation}

\newtheorem{proposition}{Proposition}
\newcommand{\mysubsection}[1]{\medskip\noindent\textbf{#1}}
\newcommand{\PComplexity}{PTIME}
\newcommand{\NPComplexity}{NP}

\newcommand{\coNPComplexity}{co-NP}

\newcommand{\StoPComplexity}{$\Sigma_{2}^{P}$} % S2P
\newcommand{\prodComplexity}{$\Pi_{2}^{P}$} % S2P

\newcommand{\countPComplexity}{\#P}

\usepackage{amssymb} % for \mathbb
\usepackage{amsmath} % for \mathbb
\usepackage[T1]{fontenc}
\newcommand{\z}{\textbf{z}}
\newcommand{\x}{\textbf{x}}
\newcommand{\y}{\textbf{y}}
\newcommand{\w}{\textbf{w}}
\newcommand{\X}{\{1,\dots,n\}}
\newcommand{\yes}{\textit{Yes}}
\newcommand{\no}{\textit{No}}

\usepackage{hyperref}
\usepackage{xspace}
\usepackage{algpseudocode}
\newcommand{\mySuff}{\text{suff}\xspace}
\newcommand{\gq}{\text{$G$-$Q$}\xspace}

\usepackage[accepted]{icml2024}

\icmltitlerunning{Local vs. Global Interpretability: A Computational Complexity Perspective}

\begin{document}

\twocolumn[
\icmltitle{Local vs. Global Interpretability: \\ A Computational Complexity Perspective}




\begin{icmlauthorlist}
\icmlauthor{Shahaf Bassan}{he}
\icmlauthor{Guy Amir}{he}
\icmlauthor{Guy Katz}{he}
\end{icmlauthorlist}

\icmlaffiliation{he}{The Hebrew University of Jerusalem, Jerusalem, Israel}
\icmlcorrespondingauthor{Shahaf Bassan}{shahaf.bassan@mail.huji.ac.il}

\icmlkeywords{interpretability, explainable AI, XAI, formal XAI, theoretical XAI, computational complexity}

\vskip 0.3in]



\printAffiliationsAndNotice{} % otherwise use the standard text.

\begin{abstract}
The local and global interpretability of various ML models has been
studied extensively in recent years. However, despite significant
progress in the field, many known results remain informal or lack sufficient mathematical rigor. We propose a framework for bridging this gap, by using computational complexity theory to assess local and global perspectives of interpreting ML models. We begin by proposing proofs for two novel insights that are essential for our analysis:
	\begin{inparaenum}[(i)]
		\item a duality between local and global forms of
		explanations; and
		\item the inherent uniqueness of certain global explanation forms.
	\end{inparaenum}
	We then use these insights to evaluate the complexity of computing explanations, across three
	model types representing the extremes of the
	interpretability spectrum:
	\begin{inparaenum}[(i)]
		\item linear models;
		\item decision trees; and 
		\item neural networks.
	\end{inparaenum}
Our findings offer insights into both the local and global
interpretability of these models. For instance, under standard
complexity assumptions such as P $\neq$ NP, we prove that selecting
\emph{global} sufficient subsets in linear models is computationally
harder than selecting \emph{local} subsets. Interestingly, with neural
networks and decision trees, the opposite is true: it is harder to carry out this task locally than globally. %A similar pattern is also observed in the task of identifying redundant features.
%These findings emphasize the importance of examining explainability through a computational complexity lens to develop a more rigorous grasp of the inherent interpretability of ML models.
We believe that our findings demonstrate how examining explainability through a computational complexity lens can help us develop a more rigorous grasp of the inherent interpretability of ML models.
\end{abstract}

\section{Introduction}
% \guy{I think the ``math commands'' file is not up-to-date with their
	%	2024 files. I tried updating it, but it breaks some commands. I
	%	think this is not crucial, and we can keep the current version}
%Global interpretability is the ability to understand the overall decision logic of an ML model. Local interpretability, on the other hand, is the ability to trace back and understand specific decisions made by that model~(\cite{zhang2021survey, du2019techniques}). Various claims have been made concerning the global and local interpretability of different ML models~(\cite{gilpin2018explaining}). For example,~\cite{molnar2020interpretable} argues that linear classifiers possess inherent local interpretability, but lack inherent global interpretability. Although linear classifiers seem like they could be perfectly globally interpreted by their weights, ~\cite{molnar2020interpretable} argues that the interpretation of a weight is interlocked with all other weights, and thus only makes sense within the context of the feature values being fixed. Decision trees, however, are considered to be both locally and globally interpretable, since their general structure is transparent and it is possible to trace the path of every decision (\cite{gilpin2018explaining, du2019techniques}). Nonetheless, if a tree becomes significantly deep, it can become substantially harder to understand its underlying decision rule~(\cite{molnar2020interpretable, izza2022tackling}). 

Interpretability is becoming an increasingly important aspect
of ML models, as it plays a key role in ensuring their safety,
transparency and fairness~\cite{doshi2017towards}. The ML
community has been studying two notions of interpretability:
\emph{global interpretability}, aimed at understanding the overall
decision logic of an ML model; and \emph{local interpretability},
aimed at understanding specific decisions made by that
model~\cite{zhang2021survey, du2019techniques}. The correlation
between a model's local and global interpretability levels is not
always entirely evident. For instance, ~\cite{molnar2020interpretable}
argues that while the weights linked to a linear classifier
can be used in interpreting its local decisions, this  may not be the
case for its global behavior. 

%Although linear classifiers seem like they could be perfectly globally interpreted by their weights, ~\cite{molnar2020interpretable} argues that the interpretation of a weight is interlocked with all other weights, and thus only makes sense within the context of the feature values being fixed. Decision trees, however, are considered to be both locally and globally interpretable, since their general structure is transparent and it is possible to trace the path of every decision (\cite{gilpin2018explaining, du2019techniques}). Nonetheless, if a tree becomes significantly deep, it can become substantially harder to understand its underlying decision rule~(\cite{molnar2020interpretable, izza2022tackling}). 

Despite significant progress in ML interpretability techniques, there still remains a notable lack of mathematical rigor in our comprehension of the inherent interpretability of different ML models. The work of~\cite{BaMoPeSu20} proposes addressing
this gap by analyzing interpretability through the perspective of
\emph{computational complexity} theory. There, the goal is to
deepen our understanding of interpretability by 
 exploring the computational complexity involved in generating
 different kinds of explanations for various ML models. A model is
 considered interpretable if an explanation can be computed
 efficiently; and conversely, if deriving an explanation is computationally intractable, the model is regarded as uninterpretable.

The study on the complexity of obtaining explanations includes various
ML models and diverse forms of
explanations~\cite{marques2020explaining, ArBaBaPeSu21,
  arenas2022computing, waldchen2021computational,
  marques2021explanations}. However, this prior work focused mainly on
\emph{local} forms of explanations --- enabling the formal analysis of \emph{local} interpretability across various
contexts, rather than addressing the overarching global interpretability of these models.

%Here, we argue that a more formal basis is required in order to
%establish a sound theoretical foundation for assessing
%interpretability. To do so, we propose to study interpretability
%through the lens of \emph{computational complexity} theory. We seek to
%study different notions of model explanations, and determine the
%computational complexity of obtaining them --- in order to measure the
%interpretability of different models. Intuitively, the lower the
%complexity, the more interpretable the model is with respect to that
%form of explanation. Recent work provided new insights in this
%direction~\cite{BaMoPeSu20, marques2020explaining,
%	arenas2022computing}, but focused mainly on \emph{local} forms of
%explanations --- thus contributing to
%formally measuring the \emph{local} interpretability across various
%contexts, rather than addressing the issue of global interpretability.

%However, this work has focused primarily on
%\emph{local} forms of explanations, which provide insights for
%predictions over specific instances. %In contrast, global
%interpretability, which may serve to assess a model's explainability
%more broadly, and in a way that does not depend on particular inputs,
%has received little attention.
\textbf{Our contributions.} We present a formal computational-complexity-based framework for
evaluating both the local and global interpretability levels of ML models. We do this by analyzing the complexity associated with computing different forms of explanation, distinguishing between those that are local (specific to a particular instance $\x$) and global (applicable to any potential instance $\x$).

\begin{figure*}[h]
	\centering
	\hspace*{-1cm}\includegraphics[width=0.6\textwidth]{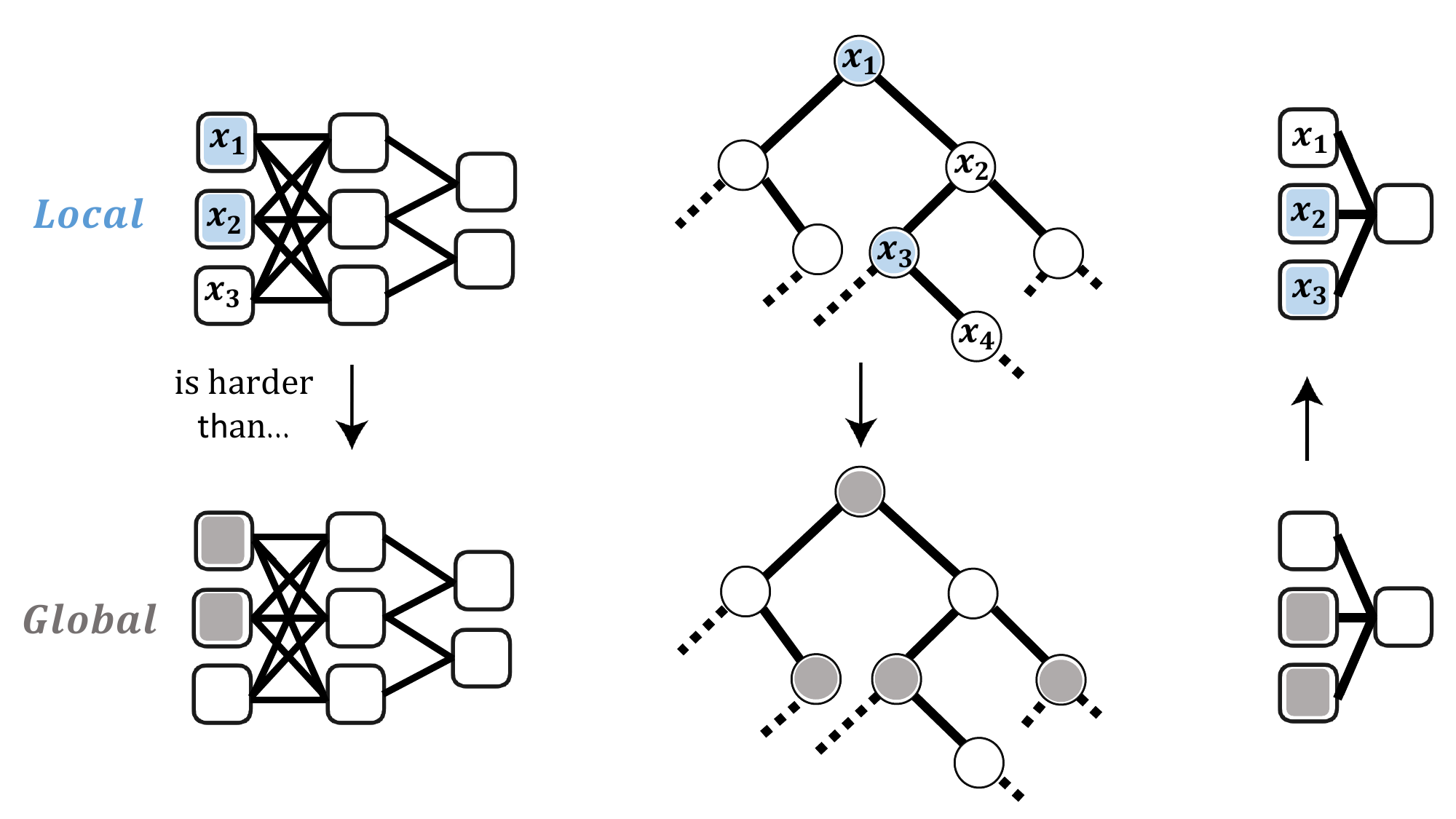}
	\caption{Illustration of complexity separations between local and global explanations. In linear models, it is harder to identify the smallest \emph{global} sufficient subset (highlighted in gray) compared to a local one (highlighted in blue). Interestingly, this reverses in neural networks and decision trees, where selecting the smallest \emph{global} sufficient subsets is computationally simpler than finding the smallest \emph{local} ones. %Global sufficient subsets are marked in gray, while local sufficient subsets are highlighted in blue.}
	}\label{fig:pdfplot}
\end{figure*}

Our study focuses on the analysis of \emph{formal} notions of
explanations that satisfy logical and mathematical guarantees ---
a sub-field often referred to as  \emph{formal explainable
  AI}~\cite{marques2022delivering}. The ability to deliver
explanations with mathematically provable guarantees is crucial in
safety-critical systems, and it also makes it possible to rigorously
assess the computational complexity of obtaining such explanations. In this
context, we focus on a few, commonly used formal notions of explanations:

%To do this, %we present a formal framework for
%evaluating the local and global interpretability of ML models. The
%we provide a formal framework that rigorously assesses the computational complexity required to
%obtain various explanation forms, either local (pertaining to a
%specific instance $\x$) or global (pertaining to any possible
%$\x$). Consequently, it affords insights into both the local
%and global interpretability level of the various models and
%explanation forms. We focus on explanation forms that are relevant to the following settings: 

%We focus on the following forms of explanations, whose local
%forms were studied previously:
% starts off with an examination of various types of
% explanations previously explored in the literature. While prior
% theoretical frameworks predominantly concentrated on local forms of
% these explanations, our work encompasses both local and global
% variations. In particular, we focus on the following forms of
% explanations:
% These include sufficient reasons, i.e., subsets of features that suffice the verdict, validating whether features are necessary or redundant to determine a prediction, as well as counting the portion of completions of a prediction, which relates to the the probability of obtaining that prediction.

	% \item \textbf{Checking Sufficient reasons.} Determine whether a subset of features is a \emph{sufficient reason}. Whereas global sufficient reasons imply that this subset always suffices the verdict, local sufficient reasons imply that this subset suffices the verdict under a partial assignment of some input.

\begin{enumerate}
    \item \textbf{Sufficient Reason Feature Selection.} In the feature selection setting, users typically choose the $k$ most significant input features. This
    selection can be executed either locally (selecting
    features that influence a particular prediction) or globally
    (selecting features that affect all instances within the
    domain). We examine the widely recognized \emph{sufficiency}
    criterion, and explore the complexity of selecting subsets of
    features with the \emph{smallest possible cardinality} while still
    maintaining sufficiency.
  \item \textbf{Necessary and Redundant Features.}
We analyze the computational complexity involved in identifying
features that are either highly important or highly redundant. This
type of analysis can be carried out in either a local or a global setting.
    \item \textbf{Completion Count.} We consider a relaxed version of the former explainability forms, which computes the \emph{relative portion} of assignments that maintain a prediction, given that we fix some subset of features. This form relates to the \emph{probability} of obtaining a prediction, and can also be computed either locally or globally.
\end{enumerate}

While the complexity of some of the \emph{local} variants of these explanation
forms has been studied previously, we focus here on their global
variants. 
As part of our analysis, we present two novel theoretical insights: % In order to fully understand the complexity analysis of our work, we start by establishing two novel theoretical insights, which will later have a crucial impact on our assessments:
\begin{inparaenum}[(i)]
	\item a \emph{duality} between local and global forms
	of explanations; and
	\item a result on the \emph{uniqueness} of global sufficiency-based explanations, in stark contrast to the \emph{exponential
	abundance} of their local counterparts.
\end{inparaenum}
Using these insights, we are able to establish hitherto unknown
complexity results on three model types that
are frequently mentioned in the literature as being at the extremities of
the interpretability spectrum:
\begin{inparaenum}[(i)]
	\item decision trees;
	\item linear models; and 
	\item neural networks.
\end{inparaenum}

In some cases, our complexity results rigorously justify prior
claims. For example, we establish that linear models are indeed easier
to interpret locally than globally under some
contexts~\cite{molnar2020interpretable} ---
selecting local sufficient subsets in these models can be
performed in polynomial time, but selecting global
sufficient subsets is coNP-Complete.

In other cases, however, our results actually defy intuition. For
example, we discover that selecting global sufficient subsets is more tractable
than local sufficient subsets, both for neural networks and decision
trees (see illustration in figure~\ref{fig:pdfplot}). In the case of decision trees, for instance, the global form of
this task can be performed in polynomial time, but the task becomes
NP-Complete for its local counterpart. A similar phenomenon occurs
with respect to identifying redundant features --- this task is 
computationally harder to perform locally than globally,
both for neural networks and decision trees.

We believe that these findings underscore the importance of rigorously
analyzing the complexity of obtaining explanations, in order to enhance our understanding of model interpretability. While our study, like others in this field, is constrained by the specific explanation forms evaluated, we believe it provides a solid basis for deeper insights into both local and global interpretability, and paves the way for future exploration of additional explanation forms.
% A comprehensive analysis of all relevant complexity classes is available in Section~\ref{}.

Due to space limitations, we provide a brief outline of the proofs for some of our claims within the paper, while the full proofs for all claims are relegated to the appendix.

\section{Preliminaries}

%\subsection{Domain}
% We assume a set of $n$ categorical features $X=\{x_1,\ldots,x_n\}$,
% % \guyamir{how can they be ordinal?}
% % (categorical or ordinal)
% where the domains of each feature are set to be $y_i= \{0,1\}$. The entire feature space will be denoted as $\mathbb{F}=y_1\times \ldots \times y_n$.
% We seek to locally interpret the prediction of a binary classifier $f:\mathbb{F}\to \{0,1\}$, i.e., given some $x\in \mathbb{F}$, to locally explain the prediction $f(x)$. %In many cases we would want to assess explanations, with respect to some distribution and disregard any OOD assignment that may cause the explanation to result in irrelevant behavior. Since we base our work on previous formalizations of binary classification, the corresponding distribution over which the explanation is provided is only meaningful in order to indicate whether a given input is in-distribution (ID) or OOD.
% Since in this sense, we are only interested in a binary indicator that evaluates whether a given input is in-distribution or out-of-distribution,
% todo{remove}
%We define a \emph{distribution-indicator} $d$ as $d(x)=\mathbbm{1}_{\{x \ is \ ID\}}$. A distribution-indicator clearly satisfies all the information that is needed in order to determine whether a certain instance is ID or OOD.

% \todo{EG: I would suggest the following. If you don't like it, delete this paragraph and for $\x$ just change in the "newcommand" section at the top of the latex doc}

% The term ``local'' means that we provide an explanation over the specific input $\x$. 

%\subsection{Complexity classes and Domain}
\textbf{Complexity Classes.} The paper assumes basic familiarity with the common complexity classes
of polynomial time (PTIME) and nondeterministic polynomial time
(\NPComplexity, \coNPComplexity).
% of polynomial complexity time (\PComplexity) and nondeterministic polynomial time (\NPComplexity, \coNPComplexity).
We also  mention classes of the second order of
polynomial hierarchy, i.e., \StoPComplexity, which describes the set of
problems that could be solved in \NPComplexity{} given  an
oracle that solves problems of \coNPComplexity{} in constant time, and
\prodComplexity, which describes the set of problems that could be
solved in \coNPComplexity{} given  an oracle that solves
problems of \NPComplexity{} in constant time.  Both
\NPComplexity{} and \coNPComplexity{} are contained in both
\StoPComplexity and \prodComplexity, and it is also widely believed that
this containment is strict i.e., \PComplexity $\subsetneq$
\NPComplexity, \coNPComplexity $ \subsetneq$ \StoPComplexity,
\prodComplexity~\cite{arora2009computational}. We also
discuss the class \countPComplexity, which corresponds to the total
number of accepting paths of a polynomial-time nondeterministic Turing
machine. It is also widely believed that \countPComplexity{} strictly
contains the second order of the polynomial hierarchy, i.e., that
\StoPComplexity, \prodComplexity $\subsetneq$
\countPComplexity~\cite{arora2009computational}.

\textbf{Setting.} We assume a set of $n$ input feature assignments
$\x:=(\x_1,\ldots,\x_n)$, and use $\mathbb{F}:=\{0,1\}^n$ to denote the
entire feature space. Our goal is to interpret the prediction of a
classifier $f:\mathbb{F}\to \{0,1\}$. In the \emph{local} case, we
seek the reason behind the prediction $f(\x)$ for a specific
instance $\x$. In the \emph{global} case, we seek to explain the
general behavior of $f$. We follow common practice and use
boolean input and output values to simplify the
presentation~\cite{ArBaBaPeSu21, waldchen2021computational,
	BaMoPeSu20}. We note, however, that many of our results carry over
to the real-valued case as well; see appendix~\ref{appendix:extension_to_continuous_inputs} for additional
information.

%\subsection{Local and Global Explainability queries}

\textbf{Explainability Queries.}  To cope with the abstract nature of interpretability, prior work often uses a construct called 
an \emph{explainability query}~\cite{BaMoPeSu20},
denoted $Q$, which defines an explanation of a specific type.
%These queries are used to provide a reasoning behind a model and typically provide answers to decision problems or to counting problems. 
As prior work focused mainly on \emph{local} explanation forms, the
input of $Q$ is usually comprised of both $f$ and a specific $\x$, and its output
is an answer providing information regarding the interpretation of
$f(\x)$. For any given explainability query $Q$, we define its
corresponding \emph{global form of explanation} as $\gq$. In contrast
to $Q$, the input of $\gq$ does not include a specific instance $\x$,
and the output conditions hold for \emph{any} $\x\in \mathbb{F}$.
We provide the full formalization of each
local and global explainability query in
Section~\ref{forms_of_explanations}.

%We acknowledge that in many cases explanations are  forms of explanations are obtained for any input in some particular domain $D\in \mathbb{F}$ and not for any instance in the 

%\subsection{Models.}
%\textbf{Classes of Models.} While sections~\ref{forms_of_explanations}
%and~\ref{properties_of_global_explanations} are general, we dive into
%the analysis of specific model types on
%section~\ref{complexitysection1}. We focus here on analyzing linear
%models, decision trees, and neural networks. More specifically, we
%address Free binary decision diagrams (FBDDs), which are a
%generalization of decision trees, Perceptrons, and Multi-layer
%perceptrons (MLPs) with reLU activation units.

\section{Local and Global Explanation Forms}
\label{forms_of_explanations}

Although model interpretability is subjective, there are several
commonly used notions of local and global explanations, on which we
focus here:

% Although model interpretability is subjective, certain established
% practices offer a solid foundation for formally interpreting a model's
% decisions. In our study, we explore several of these practices and
% give particular attention to both local and global explanations.

\textbf{Sufficient Reason Feature Selection.} In the feature selection setting, it is common for users to choose the top $k$ features participating in a model's decision. We consider the widely recognized \emph{sufficeincy} criterion for this selection, which aligns with common explainability methods~\cite{ribeiro2018anchors, carter2019made, ignatiev2019abduction}. We follow common conventions and define a \emph{local sufficient reason} as
a subset of features, $S\subseteq \X$, such that when features in $S$ are fixed to their corresponding values in $\x$,
the
prediction is determined to be $f(\x)$, regardless of other features'
assignments. %This definiton is quite common in previously suggested interpretability formalizations~\cite{}.
% Generally, a sufficient reason is defines as aa set of features uch that fixing the values of a sufficient reason $S$ determines that for any values set to the completion $\Bar{S}$ it holds that the classification remains $f(x)$:
Formally, S is a local sufficient reason with respect
to $\langle f,\x\rangle$ iff it holds that: 
%for any $\z\in\mathbb{F}$ it holds that
%$f(\x_{S};\z_{\Bar{S}})=f(\x)$.

\begin{equation}
	\forall(\z\in \mathbb{F}).\quad [f(\x_{S};\z_{\Bar{S}})=f(\x)]
\end{equation}

Here, $(\mathbf{x}_S;\mathbf{z}_{\Bar{S}})$
represents an assignment where the values of elements of $S$ are taken from
$\mathbf{x}$, and the remaining values $\overline{S}$ are taken
from $\mathbf{z}$.

In the \emph{global feature selection} setting, it is common to choose the top features contributing to \emph{all} instances~\cite{wang2015feature}. We define a set $S\subseteq \X$ as a \emph{global sufficient reason} of $f$ if
it is a local sufficient reason for all $\x$. More formally:

\begin{equation}
	\forall(\x,\z\in \mathbb{F}).\quad [f(\x_{S};\z_{\Bar{S}})=f(\x)]
\end{equation}

%for any
%$\x,\z \in \mathbb{F}$, it holds that %$f(\x_{S};\z_{\Bar{S}})=f(\x)$.
We denote $\mySuff(f,\x,S)=1$ when $S$ is a local sufficient reason of
$\langle f,\x\rangle$, and $\mySuff(f,\x,S)=0$ otherwise. Similarly,
we denote $\mySuff(f,S)=1$ when $S$ is a \emph{global} sufficient
reason of $f$, and $\mySuff(f,S)=0$ otherwise.
%\begin{equation}
%	\label{explanation_definition}
%	\forall(\z\in \mathbb{F}).\quad [f(\x_{S};\z_{\Bar{S}})=f(\x)]
%\end{equation}
% \guyamir{the the F values are VERY confusing, lets keep one of them (feature space) and change the symbol for the other (input space)}

A common notion in the literature suggests that smaller sufficient
reasons (i.e., with smaller $|S|$), whether local or global, are more meaningful than larger
ones~\cite{ribeiro2018anchors,
carter2019made,ignatiev2019abduction, halpern2005causes}. Consequently, it is common to consider the complexity of obtaining subsets of features of \emph{minimal cardinality} (also known as \emph{minimum sufficient reasons}). This leads us to our first explainability query:

\noindent\fbox{%
	\parbox{\columnwidth}{%
		\mysubsection{MSR (Minimum Sufficient Reason)}:
		
		\textbf{Input}: Model $f$, input $\x$, and integer k
		
		\textbf{Output}: 
		\yes{} if there exists some $S$ such that $\mySuff(f,\x,S)=1$ and $|S|\leq k$, and \no{} otherwise
	}%
}   

To differentiate between the local and global setting, we use
\emph{G-MSR} to refer to the explainability
query that obtains a cardinally minimal \emph{global} sufficient reason of
$f$. Due to space limitations, we relegate the full formalization of
global queries to appendix~\ref{global_forms_section_appendix}.

To better understand the complexity of the \emph{MSR} and \emph{G-MSR} queries, we also consider the analysis of a \emph{refined} version of this query, which instead of obtaining a cardinally minimal sufficient reason, is concerned with simply checking whether a subset of features is a sufficient reason:

%\begin{equation}
%	\label{explanation_definition}
%	\forall(\x,\z\in \mathbb{F}).\quad [f(\x_{S};\z_{\Bar{S}})=f(\x)]
%\end{equation}

\noindent\fbox{%
	\parbox{\columnwidth}{%
		\mysubsection{\textbf{CSR} (Check Sufficient Reason)}:
		
		\textbf{Input}: Model $f$, input $\x$, and subset of features $S$
		
		\textbf{Output}: \yes{} if $\mySuff(f,\x,S)=1$, and \no{} otherwise
	}%
}

Similarly, \emph{G-CSR} denotes the explainability query for checking
whether a subset of features is a \emph{global} sufficient reason. This formalization (along with all other global queries in this section) appears in appendix~\ref{global_forms_section_appendix}.

\textbf{Identifying Necessary and Redundant Features.} When interpreting a model, it is
common to measure the importance of each feature to a prediction. For a better understanding of the complexity of local and global computations, we consider here the complexity of identifying the two extreme cases:
features that are either \emph{necessary} or \emph{redundant} to a
prediction. We use the formal notation of necessity and redundancy
proposed by~\cite{huang2023feature}; and note that this notation also
aligns with other formal frameworks that deal with bias detection and
fairness~\cite{ArBaBaPeSu21, darwiche2020reasons,
  ignatiev2020towards}. There, necessary features can be
regarded as biased features, whereas redundant features are protected
features, which should not be used for decision making --- such as gender,
age, etc. (see appendix~\ref{Terminology_section_appendix} for more information). % There of course exist many other forms of
% fairness~\cite{mittelstadt2019explaining}, and this notion should be
% taken only as an example

Formally, we define feature $i$ as \emph{locally necessary} for
$\langle f,\x\rangle$ if it is contained in \emph{all} sufficient
reasons of $\langle f,\x\rangle$. Equivalently, removing $i$ from any sufficient
reason $S$ causes it to cease being sufficient; i.e., for any
$S\subseteq \X$ it holds that
$\mySuff(f,\x,S)=1\to \mySuff(f,\x,S\setminus \{i\})=0$.

%\begin{equation}
%	\label{explanation_definition}
%	\forall S\in (1,\ldots,n).\quad \mySuff(f,\x,S)=1\to \mySuff(f,\x,S\setminus i)=0
%\end{equation}

In the global case, we seek to determine whether $i$ is \emph{globally necessary} to
$f$, meaning it is necessary to \emph{all} instances of
$\langle f,\x\rangle$. Formally, for any $\x\in\mathbb{F}$ and for any
$S\subseteq \X$ it holds that
$\mySuff(f,\x,S)=1\to \mySuff(f,\x,S\setminus \{i\})=0$.

%\begin{equation}
%	\label{explanation_definition}
%	\forall \x\in\mathbb{F}. \quad \forall S\in (1,\ldots,n).\quad \mySuff(f,\x,S)=1\to \mySuff(f,\x,S\setminus i)=0
%\end{equation}

\noindent\fbox{%
	\parbox{\columnwidth}{%
		\mysubsection{FN (Feature Necessity)}:
		
		\textbf{Input}: Model $f$, input $\x$, and integer $i$
		
		\textbf{Output}: 
		\yes{} if $i$ is necessary with respect to $\langle f,\x\rangle$, and \no{} otherwise
	}%
}   

%The \emph{G-FN} formalization (along with other global queries in this section) appears in the appendix.

%Conversely, we define a feature $i$ as \emph{locally redundant} regarding $\langle f,\x\rangle$ if removing it from any sufficient
%reason $S$ does not affect $S$'s sufficiency. 
Conversely, a feature $i$ is termed \emph{locally redundant} regarding $\langle f,\x\rangle$ if its removal from any sufficient reason $S$ does not change $S$'s sufficiency. Formally, for any
$S\subseteq \X$ it holds that
$\mySuff(f,\x,S)=1\to \mySuff(f,\x,S\setminus \{i\})=1$. This is equivalent to $i$ not being contained in any \emph{minimal} sufficient reason.

%\begin{equation}
%	\label{explanation_definition}
%	\forall \x\in\mathbb{F}. \quad \forall S\in (1,\ldots,n).\quad \mySuff(f,\x,S)=1\to \mySuff(f,\x,S\setminus i)=1
%\end{equation}

\noindent\fbox{%
	\parbox{\columnwidth}{%
		\mysubsection{FR (Feature Redundancy)}:
		
		\textbf{Input}: Model $f$, input $\x$, and integer $i$.
		
		\textbf{Output}: 
		\yes{}, if $i$ is redundant with respect to $\langle f,\x\rangle$, and \no{} otherwise.
	}%
}

We say that a feature is \emph{globally redundant} if it is locally
redundant with respect to all inputs; i.e., for any $\x\in\mathbb{F}$
and $S\subseteq \X$ it holds that
$\mySuff(f,\x,S)=1\to \mySuff(f,\x,S\setminus \{i\})=1$. 

\textbf{Counting completions.}
Lastly, we explore a \emph{relaxed} version of the previous forms,
which is commonly analyzed in other formal
frameworks~\cite{BaMoPeSu20, waldchen2021computational,
  izza2021efficient}. This explanation form is based on exploring the relative
portion of assignment completions that maintain
a specific classification. This relates to the \emph{probability} that a prediction remains the same, assuming the other features are uniformly and independently distributed. We define the local completion count $c$ of $S$ as the
relative portion of completions which maintain the prediction of
$f(\x)$:
\begin{equation}
	\label{explanation_definition}
	c(S,f,\x):= \frac{{|\{\z\in \{0,1\}^{|\overline{S}|}, f(\x_{S};\z_{\Bar{S}}) = f(\x)\}|}}{{|\{\z\in \{0,1\}^{|\overline{S}|}|}}
\end{equation}
In the global completion count case, we count the number of
completions for all possible assignments $\x\in\mathbb{F}$:
\begin{equation}
	%	\label{explanation_definition}
	c(S,f):= \frac{{|\{\x\in \mathbb{F},\z\in \{0,1\}^{|\overline{S}|},  f(\x_{S};\z_{\Bar{S}}) = f(\x)\}|}}{{|\{\x\in \mathbb{F},\z\in \{0,1\}^{|\overline{S}|}}|}
\end{equation}

\noindent\fbox{%
	\parbox{\columnwidth}{%
		\mysubsection{CC (Count Completions)}:
		
		\textbf{Input}: Model f, input x, and subset of features S
		
		\textbf{Output}: The completion count $c(S,f,\x)$
	}%
}

We acknowledge that other explanation forms can be used, and do
not argue that one form is superior to others; rather, our goal is to
study some local and global versions of common explanation forms as a
means of assessing the local and global interpretability of different
ML models.

% We acknowledge the abstract nature of model interpretability, where
% additional forms of explanations can be proposed. We do not by any
% means assert the existence of a single ``correct'' form of
% explanation. Instead, we illustrate how various interpretability
% forms, that were previously studied, can be evaluated with respect to
% both there local and global
% configurations. %In the upcoming section, we delve into the formalization of a more comprehensive set of explainability queries that encompass the fundamental aspects of all the aforementioned explanation types.
% % , in a given context.
% % all explanation forms that were discussed in this section can be formalized as a more general set of explainability queries.

\section{Properties of Global Explanations}
\label{properties_of_global_explanations}

We now present several novel results concerning the characteristics of
the aforementioned local and global forms of explanation. Subsequently,
 in Section~\ref{complexitysection1} we
 illustrate how these results significantly affect the
complexity of computing such explanations.

% While the previous section mentioned sufficient reasons as a form of explanation, sometimes the dual . A contrastive reason is a subset of features, the alternation of which may cause the misclassification of $f(\x)$. Formally put:

% \begin{equation}
	% \label{explanation_definition}
	% \exists \z\in \mathbb{F}.\quad [f(\x_{S};\z_{\Bar{S}})\neq f(\x)]
	% \end{equation}

% Contrastive reasons are also well related to necessity. It is striaghtforward to prove the following claim (proof is attached in the appendix):

% \begin{theorem}
	% A feature $i$ is necessary w.r.t $\langle f,\x\rangle$ if and only if $\{i\}$ is a contrastive reason of $\langle f,\x\rangle$.
	% \end{theorem}

\subsection{Duality of Local and Global Explanations}

Our analysis shows that there exists a dual relationship between local
and global explanations. To better understand this relationship,
we make use of the definition of \emph{contrastive reasons},
which describes subsets of features that, when altered, may cause the
classification to change. Formally, a subset of features $S$ is a
contrastive reason with respect to $\langle f,\x\rangle$ iff there exists some $\z\in\mathbb{F}$ such that $f(\x_{\Bar{S}};\z_{S})\neq f(\x)$.

%\begin{equation}
%	\label{explanation_definition}
%	\exists \z\in \mathbb{F}.\quad [f(\x_{S};\z_{\Bar{S}})\neq f(\x)]
%\end{equation}

While sufficient reasons provide answers to ``\emph{why?}'' questions,
i.e., ``why was $f(\x)$ classified to class $i$?'', contrastive reasons
seek to provide answers to ``\emph{why not?}'' questions. Clearly, $S$ is
a sufficient reason of $\langle f,\x\rangle$ iff $\overline{S}$ is \emph{not} a
contrastive reason of $\langle f,\x\rangle$. Contrastive reasons are also well
related to necessity. This is shown by the following theorem,
whose proof appears in appendix~\ref{duality_proofs_appendix}: 

\begin{theorem}
	\label{necessetiy_contrastive_theorem}
	A feature $i$ is necessary with respect to $\langle
	f,\x\rangle$ if and only if $\{i\}$ is a contrastive reason of
	$\langle f,\x\rangle$.
\end{theorem}

We can similarly define a \emph{global contrastive reason} as a subset
of features that may cause a misclassification for any possible
input. Formally, for any $\x\in\mathbb{F}$ there exists some
$\z\in\mathbb{F}$ such that $f(\x_{\Bar{S}};\z_{S})\neq f(\x)$. This
leads to a first dual relationship between local and global
explanations:

%\begin{equation}
%	\label{explanation_definition} \forall \x,\exists
%	\z\in \mathbb{F}.\quad [f(\x_{S};\z_{\Bar{S}})\neq f(\x)]
%\end{equation}

\begin{theorem}
	\label{duality_one} Any global sufficient reason of $f$ intersects
	with all local contrastive reasons of $\langle f,\x\rangle$, and any
	global contrastive reason of $f$ intersects with all local
	sufficient reasons of $\langle f,\x \rangle$.
\end{theorem}

This formulation can alternatively be expressed through the concept of
\emph{hitting sets} (additional details appear in appendix~\ref{duality_proofs_appendix}). In
this context, global sufficient reasons correspond to hitting sets of
local contrastive reasons, while local contrastive reasons correspond
to hitting sets for global sufficient reasons. It follows that the
minimum hitting set (MHS; see appendix~\ref{duality_proofs_appendix}) aligns with cardinally minimal
reasons. Formally:

\begin{theorem}
	The MHS of all local contrastive reasons of $\langle f,\x\rangle$ is
	a cardinally minimal global sufficient reason of $f$, and the MHS of
	all local sufficient reasons of $\langle f,\x\rangle$ is a
	cardinally minimal global contrastive reason of $f$.
\end{theorem}

For instance, suppose the set of all local contrastive reasons of $f$ is $\mathbb{C}:=\{\{1,2\}, \{2,3,4\},\{4,5,6\}\}$ (these may correspond to different local inputs $\x\in\mathbb{F}$). The smallest set intersecting all subsets in $\mathbb{C}$ (hence representing its MHS) is $\{2,4\}$, thus $\{2,4\}$ is the cardinally minimal \emph{global} sufficient reason for $f$. Conversely, if $\mathbb{C}$ represents all local sufficient reasons, $\{2,4\}$ represents the cardinally minimal global contrastive reason for $f$.

\subsection{Uniquness of Global Explanations}

As stated earlier, small sufficient reasons are often assumed to
provide a better interpretation than larger ones. Consequently, we are
interested in \emph{minimal} sufficient reasons, i.e., explanation
sets that cease to be sufficient reasons as soon as even one feature
is removed from them.  We note that \emph{minimal} sufficient reasons
are not necessarily cardinally minimal, and we can also consider
\emph{subset minimal} sufficient reasons (alternatively referred to as
locally minimal). The choice of the terms \emph{cardinally minimal}
and \emph{subset minimal} is deliberate, to reduce confusion with the
concepts of global and local explanations.

A greedy approach for computing subset minimal sufficient reasons appears in
Algorithm~\ref{alg:subset-minimal-local} (similar schemes appear in~\cite{ignatiev2019abduction} and~\cite{bassan2023towards}).  It starts with the entire
set of features, and then gradually attempts to remove features until
converging to a \emph{subset minimal} sufficient reason. Notably, the
validation step at Line~\ref{lst:line:sufficientline} within the
algorithm, which determines the sufficiency of a feature subset, is
not straightforward. In Section~\ref{complexitysection1}, we delve
into a detailed discussion of the computational complexities
associated with this process.

\begin{algorithm}
	\algnewcommand\algorithmicforeach{\textbf{for each}}
	\algdef{S}[FOR]{ForEach}[1]{\algorithmicforeach\ #1\ \algorithmicdo}
	\textbf{Input} $f$, $\x$
	\caption{Local Subset Minimal Sufficient Reason}\label{alg:subset-minimal-local}
	\begin{algorithmic}[1]
		\State $S \gets \{1,\ldots,n\}$
		\ForEach {$i \in \{1,...,n\}$ by some arbitrary ordering}\label{lst:line:orderinglinelocal}
		\If{$\mySuff(f,\x,S\setminus \{i\})=1$}\label{lst:line:sufficientline}
		\State $S \gets S\setminus \{i\}$
		\EndIf
		\EndFor\label{lst:line:endregularupper}
		\State \Return $S$ \Comment{$S$ is a subset minimal \emph{local} sufficient reason}
	\end{algorithmic}
\end{algorithm}

While Algorithm~\ref{alg:subset-minimal-local} converges to a
subset-minimal local sufficient reason, it is not necessarily a
cardinally minimal sufficient reason. This is due to the algorithm's
strong sensitivity to the order in which we iterate over features
(Line~\ref{lst:line:orderinglinelocal}). The number of subset-minimal and
cardinally minimal sufficient reasons depends on the function $f$. Nevertheless, it can be
shown that their prevalence is, in the worst case,
\emph{exponential} in the number of features $n$:

\begin{proposition}
\label{exponential_abundance}

	There exists some $f$ and some $\x\in\mathbb{F}$ such that there are $\Theta(\frac{2^{n}}{\sqrt{n}})$ local subset
	minimal or cardinally minimal sufficient reasons of $\langle f,\x\rangle$.
      \end{proposition}

A similar, greedy approach for computing subset minimal \emph{global} sufficient reasons appears in
Algorithm~\ref{alg:subset-minimal-global}:

\begin{algorithm}
	\algnewcommand\algorithmicforeach{\textbf{for each}}
	\algdef{S}[FOR]{ForEach}[1]{\algorithmicforeach\ #1\ \algorithmicdo}\label{lst:line:orderinglineglobal}
	\textbf{Input} $f$
	\caption{Global Subset Minimal Sufficient Reason}\label{alg:subset-minimal-global}
	\begin{algorithmic}[1]
		\State $S \gets \{1,\ldots,n\}$
		\ForEach {$i \in \{1,...,n\}$ by some arbitrary ordering}\label{lst:line:orderinglineglobal}
		\If{$\mySuff(f,S\setminus \{i\})=1$}
		\State $S \gets S\setminus \{i\}$
		\EndIf
		\EndFor\label{lst:line:endregularupper}
		\State \Return $S$\Comment{$S$ is a subset minimal \emph{global} sufficient reason}
	\end{algorithmic}
\end{algorithm}

Given that the criteria for a subset of features to constitute a
\emph{global} sufficient reason are more stringent than those for the
local case, it is natural to ask whether they are also exponentially
abundant. To start addressing this question, we establish the
following proposition:

\begin{proposition}
	\label{two_opposite_globals_dont_exist}
	If $S_1$ and $S_2$ are two global sufficient reasons of some non-trivial function $f$, then
	$S_1\cap S_2 =S \neq \emptyset$, and $S$ is a global sufficient
	reason of $f$.
\end{proposition}

%\emph{Proof Sketch.} Using the duality property of
%Theorem~\ref{duality_one}, it is possible to conclude that
%$S_1\cap S_2\neq \emptyset$. For the second part of the claim,
%whenever $S_1\subseteq S_2$ or $S_2\subseteq S_1$, the proof is
%trivial.  When that is not the case, we observe that
%$S=S_1\cap S_2$ must be a local sufficient reason with respect to
%$\langle f,\x\rangle$ for any $\x \in \mathbb{F}$, and is hence a
%global sufficient reason with respect to $f$.

From Proposition~\ref{two_opposite_globals_dont_exist} now stems the
following theorem:
\begin{theorem}
	\label{uniquness_sufficiency_theorem}
	There exists one unique subset-minimal global sufficient reason of
	$f$.
\end{theorem}

Thus, while the local form of explanation presents us with
a worst-case scenario of an \emph{exponential} number of minimal
explanations, the global form, on the other hand, offers only
a single, unique minimal explanation. As we demonstrate later,
this distinction causes significant differences in the complexity
of computing such explanations. We can now derive the
following corollary:

\begin{proposition}
	For any possible ordering of features in
	Line~\ref{lst:line:orderinglineglobal} of
	Algorithm~\ref{alg:subset-minimal-global},
	Algorithm~\ref{alg:subset-minimal-global} converges to the same
	global sufficient reason.
\end{proposition}

The uniqueness of global subset-minimal sufficient reasons also
carries implications for the assessment of feature necessity and
redundancy, as follows:

\begin{proposition}
	\label{necessery_redundant_in_subset}
	Let $S$ be the subset minimal global sufficient reason of
	$f$. For all $i$, $i\in S$ if and only if $i$ is locally necessary
	for some $\langle f,\x\rangle$, and $i\in \overline{S}$ if and only
	if $i$ is globally redundant for $f$.
\end{proposition}

In other words, subset $S$, which is the unique minimal global
sufficient reason of $f$, categorizes the features into two possible
sets: those \emph{necessary} to a specific instance $\x$, and those that
are \emph{globally redundant}. This fact is further exemplified by the
subsequent corollary:

\begin{proposition}
	\label{necessery_or_redundant_proposition}
	Any feature $i$ is either locally necessary for some $\langle f,\x\rangle$, or globally redundant for $f$.
\end{proposition}

The proofs for all propositions and theorems discussed in this section can be found in appendix~\ref{duality_proofs_appendix} for claims related to duality, and in appendix~\ref{uniquness_proofs} for those concerning uniqueness.

\section{The Computational Complexity of Global Interpretation}
\label{complexitysection1}
%To assess the computational complexity of a particular class of
%models, denoted by $C$, previous work~\cite{BaMoPeSu20} introduces the
%concept of \emph{c-interpretability}, which represents the
%computational challenge posed by a \emph{class} of functions. To
%exemplify, when $C_{MLP}$ denotes the class of MLPs, $CSR(C_{MLP})$
%characterizes the computational task associated with checking
%sufficient reasons on MLPs.

%For example, it was shown that CSR(MLP) is \coNPComplexity-Complete whereas CSR(Perceptrons) is in PTIME. This means that Perceptrons are strictly more c-interpretable than MLPs w.r.t CSR. However, for the CC explainability query both CC(MLP) and CC(Perceptrons) are \countPCompleteComplexity, which means that MLPs and Perceptrons are equivalentley c-interpretable w.r.t CC. 

We seek to comprehensively analyze the computational
complexity of producing local and global explanations, of the forms
discussed in Section~\ref{forms_of_explanations}. We perform this
analysis on three classes of models: free binary decision diagrams (FBDDs), which are a generalization of decision trees; Perceptrons; and Multi-Layer Perceptrons (MLPs) with ReLU activation units. A full formalization of these model classes is provided in appendix~\ref{appendix:model_types}.

%linear models, decision
%trees, and neural networks, \guy{This sentence is complex. Why not
	%	just name the kinds of models we work with? Instead of saying ``we
	%	study A, B and C'' which are actually ``D, E, F and G''} and specifically address Free binary
%decision diagrams (FBDDs), which are a generalization of decision
%trees, Perceptrons, and Multi-layer perceptrons (MLPs) with reLU
%activation units. A full formalization of these models is provided in
%the appendix.

We use $Q(\mathcal{C})$ (respectively, $\gq(\mathcal{C})$) to denote the computational
problem of solving the \emph{local} (respectively, \emph{global})
explainability query $Q$ on models of class $\mathcal{C}$. Table~\ref{table:results:VerificationResults} summarizes our results, and
indicates the complexity classes for model class and explanation type
pairs.

\begin{table*}[t]
	\centering
	\def\arraystretch{1.25}% 
	\setlength{\tabcolsep}{0.9em} % for the horizontal padding
	\caption{Complexity classes for pairs of explainability queries and model 
		classes. Cells highlighted in blue represent novel
                results, presented here; whereas the remaining results were
                already known previously.}
	
	\begin{tabular}{lccccccc}
		\\
		\toprule
		%\hline
		\multirow{3}{*}{} &
		\multicolumn{2}{c}{\textbf{FBDDs}} &
		\multicolumn{2}{c}{\textbf{MLPs}} &
		\multicolumn{2}{c}{\textbf{Perceptrons}} \\
		& {\ \ Local} & {\ Global} & {\  Local} & { Global} & {\  Local} & {\ 
			Global} \\
		\midrule
		%\hline
		\textbf{CSR} & PTIME & \cellcolor{blue!10}PTIME  & coNP-C  &   
		\cellcolor{blue!10}coNP-C& PTIME & \cellcolor{blue!10}coNP-C \\
		\textbf{MSR} & \NPComplexity-C & \cellcolor{blue!10}PTIME  &   
		\StoPComplexity-C &    \cellcolor{blue!10}coNP-C& PTIME & 
		\cellcolor{blue!10}coNP-C \\
		\textbf{CC} & PTIME & \cellcolor{blue!10}PTIME &  \#P-C  &   
		\cellcolor{blue!10}\#P-C& \#P-C & \cellcolor{blue!10}\#P-C \\
		\textbf{FR} & coNP-C & \cellcolor{blue!10}PTIME & \prodComplexity-C  & 
		\cellcolor{blue!10}coNP-C & \cellcolor{blue!10}coNP-C   & 
		\cellcolor{blue!10}coNP-C \\
		\textbf{FN} & PTIME & \cellcolor{blue!10}PTIME & 
		\cellcolor{blue!10}PTIME   & \cellcolor{blue!10}coNP-C & 
		\cellcolor{blue!10}PTIME  & \cellcolor{blue!10}PTIME \\
		%\hline
		\bottomrule
	\end{tabular}
	\label{table:results:VerificationResults}
\end{table*}

%While many of results in Table.~\ref{my_table} concerning \emph{local}
%explainability were already known, we are the first to prove all
%complexity results for the corresponding \emph{global}
%explanations. As these results show, in many cases there exists a
%\emph{strict disparity} between the computational effort required to
%compute local explanations versus global explanations. This highlights
%the importance of a rigorous study of the problem.

As these results demonstrate, there is often a \emph{strict disparity}
in computational effort between calculating local and global
explanations, emphasizing the need for distinct assessments of local
and global forms. We further study this disparity and investigate the
\emph{comparative} computational efforts required for local and global
explanations across various models and forms of explanations. This
examination enables us to address the fundamental question of whether
certain models exhibit a higher degree of interpretability at a global
level compared to their interpretability at a local level, within
different contextual scenarios.
%\begin{definition}
%Let Q be an explainability query, and C a class of models, such that Q(C) is in class $K_1$ and G-Q(C) is in class $K_2$. We say that: \begin{enumerate}
	%\item C is more locally c-interpretable w.r.t Q if $K_1\subsetneq K_2$ and G-Q(C) is hard for $K_2$.
	%\item C is more globally c-interpretable w.r.t Q if $K_2\subsetneq K_1$ and Q(C) is hard for $K_1$.
	%		\end{enumerate} 
%\end{definition}

%\begin{definition}
%Let Q be an explainability query, and C a class of models, such that Q(C) is in class $K_1$ and G-Q(C) is in class $K_2$. We say that: \begin{enumerate}
	%\item C is strictly more locally c-interpretable than globally c-interpretable w.r.t Q if $K_1\subsetneq K_2$ and G-Q(C) is hard for $K_2$.
	%\item C is strictly more globally c-interpretable than locally c-interpretable w.r.t Q if $K_2\subsetneq K_1$ and Q(C) is hard for $K_1$.
	%\item C is equivalentley locally and globally c-interpretable w.r.t Q if $K_1=K_2$ and both Q(C) and G-Q(C) are hard for $K_1$ and $K_2$.
	%		\end{enumerate} 
%\end{definition}

%For example, assuming that MSR(MLP) is \coNPComplexity-Complete and G-MSR(MLP) is in PTIME, then we say that MLPs are more locally c-interpretable w.r.t to MSR. If, however G-MSR is \StoPComplexity-Complete, we will say that MLPs are more globally c-interpretable w.r.t MSR.

%For example, assuming that MSR(MLP) is \StoPComplexity-Complete, and G-MSR(MLP) is \coNPComplexity-Complete (as we later will show is the case), we will say that MLPs are strictly more globally c-interpretable than locally c-interpretable.

\textbf{Local vs. Global Interpretability}
\label{localvsglobal}

We say that a model is more \emph{locally interpretable} for a given
explanation type if computing the local form of that explanation is
strictly easier than computing the global form, and say that it is
more \emph{globally interpretable} in the opposite case. We use the
notation of \emph{c-interpretablity} (computational
interpretability)~\cite{BaMoPeSu20} to study the disparity between
local and global computations of our analyzed query forms. More formally:

\begin{definition}
	Let $Q$ denote an explainability query and $\mathcal{C}$ a class of models,
	and suppose $Q(\mathcal{C})$ is in class $\mathcal{K}_1$ and G-$Q(\mathcal{C})$ is in class
	$\mathcal{K}_2$. Then:
 
	\begin{enumerate}
		\item $\mathcal{C}$ is strictly more locally $c$-interpretable with respect to
		Q iff $\mathcal{K}_1\subsetneq \mathcal{K}_2$ and G-$Q(\mathcal{C})$ is hard for $\mathcal{K}_2$.
		\item $\mathcal{C}$ is strictly more globally $c$-interpretable with respect
		to $Q$ iff $\mathcal{K}_2\subsetneq \mathcal{K}_1$ and $Q(\mathcal{C})$ is hard for $\mathcal{K}_1$.
	\end{enumerate} 
 
\end{definition}

We divide our discussion into scenarios where
computing an explanation is more challenging in the local setting, in
the global setting, or equally difficult in both settings.

\subsection{The Locally Interpretable Case}
\label{local_interpretable_case}

We start with the Perceptron model, where a strict complexity gap exists between local and global computations. Our findings reveal that in linear models, global feature selection is computationally harder than local feature selection. As shown in Table~\ref{table:results:VerificationResults}, there is a disparity in feature selection queries (\emph{CSR} and \emph{MSR}) between local and global forms. Local forms are achievable in polynomial time, whereas global forms are coNP-Complete, leading to our first corollary:

\begin{theorem}
	Perceptrons are strictly more locally c-interpretable with respect to CSR and MSR.  %For CC, FN and CSR obtaining explanations both locally or globally are polynomial.
\end{theorem}

The complexity difference arises from the intrinsic properties of linear models. Previous work showed that in linear models, selecting cardinally minimal \emph{local} sufficient reasons can be performed in polynomial time~\cite{BaMoPeSu20, marques2020explaining}. The feasibility of such algorithms stems from the understanding that the local value $\x_i$ of a feature $i$, along with its corresponding weight $\w_i$, can be utilized to calculate the exact contribution of feature $i$ to the local prediction $f(\x)$.

However, we are able to prove that for \emph{global} sufficient reasons, the situation is different, as the sufficiency criterion for the selection process considers all inputs $\x$, rather than a specific $\x$. This property makes the task of selecting global sufficient reasons for linear models intractable:

\begin{proposition}
For Perceptrons, solving G-CSR and G-MSR is coNP-Complete, while solving CSR and MSR can be done in polynomial time.
\end{proposition}

\emph{Proof Sketch.} We prove in appendix~\ref{csr_appendix} that for \emph{G-CSR}, membership in coNP holds from guessing certificates $\x\in\mathbb{F}$ and $\z\in\mathbb{F}$ and
validating whether $S$ is not sufficient. For \emph{G-MSR}, membership
is a consequence of Proposition~\ref{necessery_redundant_in_subset},
which shows that any feature that is contained in the subset minimal
global sufficient reason is necessary for some $\langle f,\x\rangle$,
or is globally redundant otherwise. Hence, we can guess $n$ assignments $\x^1,\ldots,\x^n$, and for each feature $i\in\{1,\ldots n\}$, validate whether $i$ is locally necessary for $\langle f,\x^i\rangle$, and whether this holds for more than $k$ features. We prove hardness with similar reductions for both \emph{G-CSR} and \emph{G-MSR} using a reduction from the $\overline{SSP}$ (subset-sum problem), which is coNP-Complete.

This result is particularly interesting since it provides evidence of a \emph{lack of global interpretability} in linear models, in contrast to their inherent local interpretability, supporting intuition raised by previous works~\cite{molnar2020interpretable}.

Another case where local computations were found to be strictly less
complex than global ones is identifying necessary features in neural networks:

\begin{theorem}
	MLPs are strictly more locally c-interpretable with respect to FN.%For CC, FN and CSR obtaining explanations both locally or globally are polynomial.
\end{theorem}

This disparity was only found in MLPs, as demonstrated in the following proposition:

\begin{proposition}
    For FBDDs and Perceptrons, FN and G-FN can be solved in polynomial time. However, for MLPs, FN can be solved in polynomial time, while solving G-FN is coNP-Complete.
\end{proposition}

\emph{Proof Sketch.} We prove in appendix~\ref{section_msr_appendix} that membership in coNP can be obtained using Theorem~\ref{necessetiy_contrastive_theorem}. We then prove hardness for MLPs by reducing from \emph{TAUT}, a classic coNP-Complete problem that checks whether a boolean formula is a tautology. For Perceptrons and FBDDs we suggest polynomial algorithms whose correctness is derived from Theorem~\ref{necessetiy_contrastive_theorem}.

\subsection{The Globally Interpretable Case}

The fact that a model is more locally interpretable may seem
intuitive. Nevertheless, it is rather surprising that, in certain
instances, we can prove that the \emph{global explanation forms are easier to compute than the local forms}. We demonstrate that this is sometimes the case both in neural networks and in decision trees. Specifically, in these models, local feature selection is computationally harder than global feature selection. This contrasts with linear models, where the local variant of feature selection was easier than the global:

\begin{theorem}
	FBDDs and MLPs are strictly more globally c-interpretable with respect to MSR. %For CC, FN and CSR obtaining explanations both locally or globally are polynomial.
\end{theorem}

Unlike linear models, decision trees and neural networks do not possess the characteristics that enable a polynomial selection of cardinally minimal \emph{local} sufficient reasons, which is NP-Complete for decision trees, and even less tractable ($\Sigma^P_2$-Complete) for neural networks~\cite{BaMoPeSu20}. Intuitively, NP-Hardness in decision trees stems from the need to explore an exponential number of possible subsets to find the smallest one. In neural networks, this complexity increases further because not only are there exponentially many subsets to consider, but verifying if one specific subset is a sufficient reason is already coNP-Complete, leading to the overall $\Sigma^P_2$ complexity.

However, we have demonstrated that both models have strictly lower complexity when addressing \emph{global} sufficient reasons, primarily due to the \emph{uniqueness} property of subset-minimal global sufficient reasons established in Theorem~\ref{uniquness_sufficiency_theorem}. This contrasts with their exponential abundance in the local version (Proposition~\ref{exponential_abundance}). We note that the complexity of checking whether a subset is sufficient is akin to local and global computations (\emph{CSR} and \emph{G-CSR}). However, when obtaining \emph{cardinally minimal} sufficient reasons (\emph{MSR}, \emph{G-MSR}) there exists a strict disparity between local and global computations since the complexity is obviously tied to the number of possible subset candidates. This difference renders the global query version simpler to compute than the local, due to the underlying uniqueness property:

\begin{proposition}
    For FBDDs, G-CSR and G-MSR can be solved in polynomial time, while MSR is NP-Complete. Moreover, in MLPs, solving G-CSR and G-MSR is coNP-Complete, while solving MSR is $\Sigma^P_2$-Complete.
\end{proposition}

\emph{Proof Sketch.} We prove these results in appendix~\ref{section_fn_appendix}. For FBDDs, we provide polynomial algorithms for the global queries, based on the observation that one can identify global sufficient reasons by iterating over pairs of leaf nodes, instead of iterating over single leaf nodes (which is how we identify local sufficient reasons). For MLPs, membership trivially holds, and the same hardness results used for Perceptrons hold here as well.

On the practical side, this finding demonstrates that obtaining cardinally 
minimal \emph{global} sufficient reasons is feasible in some cases. For 
decision trees, accomplishing this task is feasible within polynomial time. 
When it comes to neural networks, the task can be executed with a linear number 
of calls to a coNP oracle, such as neural network verification 
tools~\cite{wang2021beta, brix2023first, WuIsZeTaDaKoReAmJuBaHuLaWuZhKoKaBa24}.
 
This contrasts 
sharply with the \emph{local} variant of this problem, which is $\Sigma^P_2$ 
complete and thus necessitates an exponentially large number of verification 
queries, making this task infeasible, even with the use of neural network 
verifiers.

The phenomenon observed in the \emph{MSR} query, where global explanations are easier to compute for decision trees and neural networks, is mirrored in the feature-redundancy (\emph{FR}) query: it is computationally harder to identify locally redundant features than globally redundant ones in these models:

\begin{theorem}
	FBDDs and MLPs are strictly more globally c-interpretable with respect to FR. %For CC, FN and CSR obtaining explanations both locally or globally are polynomial.
\end{theorem}

The complexity of identifying redundant features is closely linked to that of the \emph{MSR} query. Recall that identifying a redundant feature is akin to validating whether a feature is not part of any subset minimal sufficient reason. While this is computationally challenging in the local scenario due to the exponential number of subset-minimal \emph{local} sufficient reasons, it becomes more tractable in the \emph{global} context.

%Hence, in the local case, this process is computationally hard due to the exponential abundance of \emph{local} subset-minimal sufficient reasons, but is more feasible for the \emph{global} case:

\begin{proposition}
        For FBDDs, G-FR can be solved in polynomial time, while solving FR is coNP-Complete. Moreover, in MLPs, solving G-FR is coNP-Complete, while solving FR is $\Pi^P_2$-Complete.
\end{proposition}

These results imply an intriguing observation regarding the complexity of identifying local and global necessary and redundant features in the specific case of MLPs:
\begin{observation}
	For MLPs, global necessity (G-FN) is strictly harder than local necessity (FN), whereas global redundancy (G-FR) is strictly less hard than local redundancy (FR).
\end{observation}

Another interesting insight from the previous theorems is the
comparison between MLPs and Perceptrons. Since Perceptrons are a
specific case of MLPs with one layer, analyzing the
complexity difference between them can provide insights into the
influence of hidden layers on model intricacy. Our findings indicate that while hidden layers influence the local queries we examined, they do not impact the global queries.

\begin{observation}
    Obtaining CSR, MSR, and FR is strictly harder for MLPs compared to MLPs with no hidden layers. However, this disparity does not exist for G-CSR, G-MSR, and G-FR.
\end{observation}

\subsection{The Equally Difficult Case}

Finally, we explore the cases in which local and global computations are complete for the same complexity class. First, we demonstrate that detecting redundant features at both local and global levels in Perceptrons is coNP-Complete:

%We demonstrate that detecting redundant features at both local and global levels in Perceptrons is coNP-Complete:

\begin{proposition}
    For Perceptrons, solving FR and G-FR are both coNP-Complete.
\end{proposition}

This stands in sharp contrast with decision trees and neural networks, where the global structure of the query is strictly simpler than its local counterpart. This difference is attributable, again, to the intrinsic properties of linear models, which lead to a diminished level of global interpretability, as discussed in subsection~\ref{local_interpretable_case}.

Next, we show that in \emph{all} of our analyzed models, the complexity of the count-completion (\emph{CC}) query remains the same across the local and global versions of computation:

\begin{proposition}
    For FBDDs, both CC and G-CC can be solved in polynomial time. Moreover, for Perceptrons and MLPs, solving both CC and G-CC are $\#P$-Complete.
\end{proposition}

\emph{Proof Sketch.} We prove these complexity results in appendix~\ref{section_fr_appendix}. Membership in $\#P$ is straightfowrad. For the hardness in the case of Perceptrons/MLPs, we reduce from (local) \emph{CC} of Perceptrons which is $\#P$-complete. For FBDDs, we propose a polynomial algorithm.

We recall that the \emph{CC} query is a \emph{relaxed} counting version of the \emph{CSR} query, which seeks the relative portion of a subset $S$, instead of posing a decision problem about $S$. This complexity result highlights a notable distinction between decision problems (which exhibit a complexity gap between local and global forms) and counting problems, where such a complexity gap is absent.

\section{Related Work}

Our work contributes to \emph{Formal XAI}~\cite{marques2020explaining}, which focuses on the analysis of
explanations with mathematical guarantees. Several
papers have explored the
computational complexity of obtaining such explanations~\cite{BaMoPeSu20,
waldchen2021computational,  arenas2022computing, ArBaBaPeSu21, 
arenas2021tractability, ordyniak2023parameterized, blanc2021provably, blanc2022query, van2022tractability, marques2021explanations, audemard2020tractable, huang2022tractable, audemard2021computational, izyamajo21}; however, these notable efforts primarily focused on local forms of explanations, while our framework offers a comprehensive approach for analyzing and contrasting both local and global explanations.

Certain terms used in our work have been referred to by additional names in the literature: ``sufficient
reasons'' are also known as \emph{abductive
explanations}~\cite{ignatiev2019relating}, while minimal sufficient
reasons are sometimes referred as \emph{prime implicants} in Boolean formulas~\cite{shih2018symbolic}. A notion
similar to the \emph{CC} query is the $\delta$-relevant
set~\cite{waldchen2021computational, izza2021efficient}, which asks
whether the completion count exceeds a threshold $\delta$. Similar
duality properties to the ones studied here were shown to hold
considering the relationship between local sufficient and contrastive
reasons~\cite{ignatiev2020contrastive}, and between
absolute sufficient reasons and adversarial
attacks~\cite{ignatiev2019relating}. Minimal \emph{absolute}
sufficient reasons are the smallest-sized subsets among all
possible inputs for a specific prediction and rely on partial input assignments. In our global
sufficient reason definition, we do not rely on particular inputs or partial assignments (see appendix~\ref{Terminology_section_appendix} for more details).

The necessity and redundancy queries that we discussed were studied
previously~\cite{huang2023feature} and were also explored under the context of bias detection~\cite{ArBaBaPeSu21,darwiche2020reasons, ignatiev2020towards}. We acknowledge, of course, that there exist many other notions of bias
and fairness~\cite{mehrabi2021survey}.

\section{Conclusion}
We present a theoretical framework using computational complexity theory to assess both local and global perspectives of interpreting ML models. Our work uncovers new insights, including a duality between local and global explanations and the uniqueness inherent in some global explanation forms. We then build upon these insights and propose novel proofs for complexity classes tied to various explanation forms, enabling us to \emph{formally measure interpretability} across different local and global contexts. While some of our findings justify folklore claims, others are unexpected. We believe that these discoveries illustrate the importance of applying computational complexity theory to gain a thorough understanding of the interpretability of ML models, paving the way for further research.

%potentially paving the way for further studies in this field.

%While some of our findings justify folklore claims, others may seem unexpected. We believe that these discoveries underscore the value of applying computational complexity theory to gain a thorough comprehension of the interpretability of ML models, potentially paving the way for further studies in this field.

%We apply these insights to commonly evaluated ML models, including linear models, decision trees, and neural networks. 

\section*{Impact Statement} 
While we acknowledge the potential social implications of interpretability, our 
work primarily focuses on theoretical aspects. Hence, we believe that it does 
not entail any direct social ramifications.

%We recognize that various other local and global explanation forms and different model types can be suggested. We do not assert our assessed explanation forms are the only ``right'' ones. Instead, our work should be taken as a study demonstrating how complexity theory can enhance our grasp of local and global interpretability in different ML models.

%We provide a theoretical framework aimed at evaluating the local
%and global interpretability of ML models, based on computational
%complexity theory. Our work provides novel insights into different
%global forms of explanations, such as a duality relationship between
%local and global forms of explanations, and the \emph{uniqueness}
%inherent in some global forms of explanations. Additionally, we provide
%novel proofs of complexity classes for global forms of
%explanations and show, how these complexity classes are affected by our
%insights regarding global forms of explanations. This in turn lets us
%formally measure both the local and global interpretability of various
%models, under different contextual scenarios, from the lens of
%computational complexity theory. We provide some of these insights to
%commonly assessed ML models, including linear models, decision trees
%and neural networks.

%\section{Experimental Evidence}

\section*{Acknowledgements}
This work was partially funded by the European Union 
(ERC, VeriDeL, 101112713). Views
and opinions expressed are however those of the author(s) only and do not 
necessarily reflect those of the European Union
or the European Research Council Executive Agency. Neither the European Union 
nor the granting authority can be held
responsible for them. The work of Amir was further supported by a scholarship 
from the Clore Israel Foundation.

\bibliography{references}
\bibliographystyle{icml2024}

%%%%%%%%%%%%%%%%%%%%%%%%%%%%%%%%%%%%%%%%%%%%%%%%%%%%%%%%%%%%%%%%%%%%%%%%%%%%%%%
%%%%%%%%%%%%%%%%%%%%%%%%%%%%%%%%%%%%%%%%%%%%%%%%%%%%%%%%%%%%%%%%%%%%%%%%%%%%%%%
% APPENDIX
%%%%%%%%%%%%%%%%%%%%%%%%%%%%%%%%%%%%%%%%%%%%%%%%%%%%%%%%%%%%%%%%%%%%%%%%%%%%%%%
%%%%%%%%%%%%%%%%%%%%%%%%%%%%%%%%%%%%%%%%%%%%%%%%%%%%%%%%%%%%%%%%%%%%%%%%%%%%%%%
\newpage
\appendix
\onecolumn
\appendix

\setcounter{definition}{0}
\setcounter{proposition}{0}
\setcounter{theorem}{0}
\setcounter{lemma}{0}
\begin{center}\begin{huge} Appendix\end{huge}\end{center}    
\noindent{The appendix contains formalizations and proofs that were mentioned throughout the paper:}

\newlist{MyIndentedList}{itemize}{4}
\setlist[MyIndentedList,1]{%
	label={},
	noitemsep,
	leftmargin=0pt,
}

\begin{MyIndentedList}
	
	\item \textbf{Appendix~\ref{global_forms_section_appendix}} formalizes the set of \emph{global} explainability queries.
	%, and its relevance to various explainability queries. 
	
	\item \textbf{Appendix~\ref{appendix:model_types}} formalizes the classes of models that were assessed in the paper.
	
	\item  \textbf{Appendix~\ref{duality_proofs_appendix}} 
	contains the proofs regarding the duality between local and global forms of explanations.

	\item \textbf{Appendix~\ref{uniquness_proofs}} contains the proofs concerning the inherent uniqueness of global forms of explanations.
	
	\item  \textbf{Appendix~\ref{csr_appendix}} contains the proof of 
	Proposition~\ref{proposition_csr_appendix} (Complexity of \emph{G-CSR} and 
	\emph{G-MSR} for Perceptrons)
	
	\item  \textbf{Appendix~\ref{section_msr_appendix}} contains the proof of 
	Proposition~\ref{msr_proof_appendix} (Complexity of \emph{FN} and 
	\emph{G-FN} for FBDDs, Perceptrons, and MLPs)

	\item  \textbf{Appendix~\ref{section_fn_appendix}} contains the proof of 
	Proposition~\ref{fn_complexity_appendix} (Complexity of \emph{G-CSR} and 
	\emph{G-MSR} for FBDDs and MLPs)
	
	\item  \textbf{Appendix~\ref{section_cc_appendix}} contains the proof of 
	Proposition~\ref{cc_complexity_appendix} (Complexity of \emph{G-FR} for 
	FBDDs and MLPs)

	\item  \textbf{Appendix~\ref{section_local_proofs_appendix}} contains the 
	proof of Proposition~\ref{local_complexity_appendix} (Complexity of 
	\emph{FR} and \emph{G-FR} for Perceptrons)

	\item  \textbf{Appendix~\ref{section_fr_appendix}} contains the proof of 
	Proposition~\ref{fr_proof_appendix} (Complexity of \emph{G-CC} for FBDDs, 
	Perceptrons, and MLPs)	

	\item  \textbf{Appendix~\ref{appendix:extension_to_continuous_inputs}}
	contains details on various extensions of this work.

	\item  \textbf{Appendix~\ref{Terminology_section_appendix}} provides details on the terminologies and forms of explanation that are relevant to those discussed here.

\end{MyIndentedList}

\section{Global forms of model explanations}
\label{global_forms_section_appendix}

In this section, we present the \emph{global} forms of the explainability queries previously mentioned, which were initially formulated in the paper for their local configuration.

\noindent\fbox{%
	\parbox{\textwidth}{%
		\mysubsection{\textbf{G-CSR} (\emph{Global} Check Sufficient Reason)}:
		
		\textbf{Input}: Model $f$,and subset of features $S$.
		
		\textbf{Output}: \yes{}, if $S$ is a global sufficient reason of $f$ (i.e., $\mySuff(f,S)=1$), and \no{} otherwise.
	}%
}

\noindent\fbox{%
	\parbox{\textwidth}{%
		\mysubsection{\textbf{G-MSR} (\emph{Global} Minimum Sufficient Reason)}:
		
		\textbf{Input}: Model $f$, and integer k.
		
		\textbf{Output}: 
		\yes{}, if there exists a global sufficient reason $S$ for $f$ (i.e., $\mySuff(f,S)=1$) such that $|S|\leq k$, and \no{} otherwise.
	}%
}  

\noindent\fbox{%
	\parbox{\textwidth}{%
		\mysubsection{\textbf{G-FR} (\emph{Global} Feature Redundancy)}:
		
		\textbf{Input}: Model $f$, and integer $i$.
		
		\textbf{Output}: 
		\yes{}, if $i$ is globally redundant with respect to $f$, and \no{} otherwise.
	}%
}  

\noindent\fbox{%
	\parbox{\textwidth}{%
		\mysubsection{\textbf{G-FN} (\emph{Global} Feature Necessity)}:
		
		\textbf{Input}: Model $f$, and integer $i$.
		
		\textbf{Output}: 
		\yes{}, if $i$ is globally necessary with respect to $f$, and \no{} otherwise.
	}%
}  

\noindent\fbox{%
	\parbox{\textwidth}{%
		\mysubsection{\textbf{G-CC} (\emph{Global} Count Completions)}:
		
		\textbf{Input}: Model $f$, and subset $S$.
		
		\textbf{Output}: 
		The global completion count $c(S,f)$
	}%
}

%Membership G-MSR is in NP:

\section{Model Classes}

\label{appendix:model_types}
Next, we describe in detail the various model classes that were taken into 
account within this work:

\mysubsection{Free Binary Decision Diagram (FBDD).} A \emph{BDD} is 
a graph-based model that represents a Boolean function $f: \mathbb{F}  
\to \{0,1\}$~\cite{Le59}. The arbitrary Boolean function is realized by an 
acyclic 
(directed) graph, for which the following holds: 
\begin{inparaenum}[(i)]
	\item every internal node $v$ corresponds 
	with a single binary input feature $(1,\ldots,n)$;
	\item every internal node $v$ has exactly two output edges, 
	that represent the values $\{0,1\}$ assigned to $v$;
	\item each leaf node corresponds to either a \emph{True}, or \emph{False}, label; 
	and % \todo{make nicer T/F values}
	\item every variable appears \emph{at most} once, along every path 
	$\alpha$ of the BDD.
\end{inparaenum}

Hence, any assignment to the inputs $\x\in\mathbb{F}$ corresponds to one unique path $\alpha$ from the BDD's root to one of its leaf nodes, and any path $\alpha$ corresponds to some partial assignment $\x_S$. We denote $f(\x):=1$ if the label of the leaf node is true, and $f(\x):=0$ if it is false. Moreover, we regard the size of a BDD (i.e., $\vert f \vert$ ) to be the overall number of 
edges in the BDD's graph. 
In this work, we focus on the popular variant of \emph{Free BDD} (FBDD) 
models, for which different paths, $\alpha,\alpha'$ are allowed to have 
different orderings of the input 
variables $\{1,\ldots,n \}$ and on every path $\alpha$ no two nodes have the same label.  A \emph{decision tree} can be essentially described as an FBDD whose foundational graph structure is a tree.

In our representation of the path $\alpha$, we denote the nodes participating in the path as $\{\alpha_1,\alpha_2,\ldots,\alpha_{t}\}$, where $t$ represents the total number of nodes in $\alpha$. Each node $\alpha_i$ within the path $\alpha$, is defined similarly to any general node in the FBDD. However, while any node possesses two output edges, we specify that a \emph{path} node possesses a \emph{single} \emph{path} output edge. This edge corresponds to the value assigned to the binary feature associated with node $\alpha_i$ (either $1$ or $0$) in its corresponding partial assignment.

\mysubsection{Multi-Layer Perceptron (MLP).} A Multi-Layer 
Perceptron~\cite{GaDo98, RaGhEtAm16} $f$ with $t-1$ hidden layers 
($g^{j}$ for $j\in\{1,\ldots, t-1\}$) and a single output layer 
($g^{t}$), is recursively defined as follows:
$g^{(j)} := \sigma^{(j)}(g^{(j-1)}W^{(j)} + b^{(j)}) \quad 
(j \in \{1,\ldots,t\})$, given $t$ 
weight matrices $W^{(1)},\ldots,W^{(t)}$, $t$ bias vectors 
$b^{(1)},\ldots,b^{(t)}$, and also $t$ activation functions 
$\sigma^{(1)},\ldots,\sigma^{(t)}$.

The MLP $f$ outputs the value $f \coloneqq g^{(t)}$, while 
$g^{(0)}\coloneqq \x \in \{0,1\}^n$ is the input layer that receives the input 
of the model. The biases and weight matrices are defined by a series of 
positive values $d_{0},\ldots,d_{t}$ that represent their dimensions. Furthermore, we assume that all the weights and biases possess rational values, denoted as $W^{(j)}\in \mathbb{Q}^{d_{j-1}\times d_{j}}$ and $b^{(j)}\in 
\mathbb{Q}^{d_{j}}$, which have been acquired during the training phase.
Due to our focus on \emph{binary} 
classifiers over $\{1,\ldots,n\}$, it necessarily holds that: $d_{0} = n$ 
and $d_{t} = 1$. In this work, we focus on the popular $ReLU(x)=\max(0,x)$ 
activation function, with the exception 
of the single activation in the last layer, that is typically a 
sigmoid function. Nonetheless, given our emphasis on post-hoc interpretations, 
it is without loss of generality that we may assume the last activation 
function is represented by the step function, i.e., $step(\z)=1 
\iff \z > 0$.

\mysubsection{Perceptron.} A Perceptron~\cite{RaReHe03} is a single-layered MLP 
(i.e., $t=1$): $f(\x) = step((\mathbf{w}\cdot\x)+b)$, for $b\in\mathbb{Q}$ and 
$\mathbf{w}\in\mathbb{Q}^{n\times d_{1}}$. Thus, for a Perceptron $f$ the following 
holds w.l.o.g.: $f(\x)=1 \iff (\mathbf{w}\cdot\x)+b \geq 0$.

\section{The Duality of Local and Global Explanations}
\label{duality_proofs_appendix}

\textbf{Minimum Hitting Set (MHS).} Given a collection
$\mathbb{S}$ of sets from a universe U, a hitting set $h$ for
$\mathbb{S}$ is a set such that
$\forall S \in \mathbb{S}, h\cap S \neq \emptyset$. A hitting set $h$
is said to be \emph{minimal} if none of its subsets is a hitting set,
and \emph{minimum} when it has the smallest possible cardinality among
all hitting sets.

\begin{theorem}
	\label{necessitiy_contrastive_iff}
	A feature $i$ is necessary with respect to $\langle f,\x\rangle$ if and only
	if $\{i\}$ is a contrastive reason of $\langle f,\x\rangle$.
\end{theorem}

\emph{Proof.} For the first direction, let us begin by assuming that $\{i\}$ is a contrastive reason with respect to $\langle f,\x\rangle$. It then follows, from the definition of sufficient and contrastive reasons, that $\{1,\ldots, n\}\setminus \{i\}$ is \emph{not} a sufficient reason for $\langle f,\x\rangle$. Consequently, any subset $S\subseteq \{1,\ldots,n\}\setminus \{i\}$ is also not a sufficient reason for $\langle f,\x\rangle$, which is equivalent to saying that for any $S\subseteq \{1,\ldots,n\}$ it holds that $\mySuff(f,\x,S\setminus \{i\})=0$. This is true whether $S$ is or is not a sufficient reason (and hence is particularly true for the case where it is one, i.e., $\mySuff(f,\x,S)=1$). As a direct consequence, for any $S\subseteq\X$ the following condition holds: $\mySuff(f,\x,S)=1\to \mySuff(f,\x,S\setminus \{i\})=0$.

For the second direction, let us assume that $i$ is necessary with respect to 
$\langle f,\x\rangle$, which particularly means that for all $S\subseteq 
\{1,\ldots,n\}$ it holds that $\mySuff(f,\x,S)=1\to \mySuff(f,\x,S\setminus 
\{i\})=0$. We now assume, by contradiction, that $\{i\}$ is not a contrastive 
reason for $\langle f,\x\rangle$. Therefore, it follows, from the very 
definition of sufficient and contrastive reasons, that $\X\setminus \{i\}$ 
\emph{is} a sufficient reason for $\langle f,\x\rangle$. Moreover, it clearly 
holds that the entire set $\X$ is a sufficient reason with respect to $\langle 
f,\x\rangle$, which is a property that holds for any $f$ and any $\x$ (fixing 
\emph{all} features necessarily determines that the prediction remains the 
same). Overall, we get that:

\begin{equation}
	\mySuff(f,\x,\X)=1\wedge \mySuff(f,\x,\X\setminus \{i\})=1
\end{equation}

This is in contradiction to the assumption that $i$ is necessary with respect to $\langle f,\x\rangle$.

\begin{theorem}
	\label{duality_one_appendix}
	Any global sufficient reason of
	$f$ intersects with all local contrastive reasons of $\langle
	f,\x\rangle$ and any global contrastive reason of $f$ intersects with
	all local sufficient reasons of $\langle f,\x \rangle$.
\end{theorem}

\emph{Proof.} For the first part, given some $f$ and some $\x$, let us assume, 
by contradiction, that there exists some global sufficient reason $S$ of $f$ 
and some local contrastive reason $S'$ of $\langle f,\x\rangle$ for which it 
holds that $S\cap S'=\emptyset$. Given that $S\cap S'=\emptyset$, it naturally 
follows that $S'\subseteq \overline{S}$. Taking into account that $S$ is a 
global sufficient reason of $f$, we can infer that $S$ is also a \emph{local} 
sufficient reason of $\langle f,\x\rangle$. Therefore, from the definition of 
sufficient and contrastive reasons, $\overline{S}$ does not qualify as a 
contrastive reason with respect to $\langle f,\x\rangle$, leading to the 
implication that no subset of $\overline{S}$ can be a contrastive reason 
either. This assertion, however, contradicts the previously established 
$S'\subseteq \overline{S}$.

The second part of the claim will be almost identical to the first part: given some $f$ and $\x$, we can 
again assume, by contradiction, that there exists some global 
\emph{contrastive} reason $S$ of $f$ and some local \emph{sufficient} reason 
$S'$ of $\langle f,\x\rangle$ for which it holds that: $S\cap S'=\emptyset$.  
Given that $S\cap S'=\emptyset$, it naturally follows that $S'\subseteq 
\overline{S}$. Since $S$ is a global contrastive reason of $f$ it also acts as a 
local contrastive reason for $\langle f,\x\rangle$. As a consequence, from the very definition of sufficient and contrastive reasons, $\overline{S}$ 
can not be a sufficient reason for $\langle f,\x\rangle$. This implies that no 
subset of $\overline{S}$ can serve as a sufficient reason for $\langle f,\x\rangle$, 
creating a contradiction with the premise that $S'\subseteq \overline{S}$.

\begin{theorem}
	The MHS of all local contrastive reasons of $\langle f,\x\rangle$ is
	a cardinally minimal global sufficient reason of $f$, and the MHS of
	all local sufficient reasons of $\langle f,\x\rangle$ is a
	cardinally minimal global contrastive reason of $f$.
\end{theorem}

Given some $f$, we denote $\mathbb{S}$ as the set of all local sufficient 
reasons of $\langle f,\x\rangle$ and denote $\mathbb{C}$ as the set of all 
local contrastive reasons of $\langle f,\x\rangle$. As a direct consequence of 
Theorem~\ref{duality_one_appendix}, we can determine the following claim:

\begin{lemma}
	\label{duality_not_minimum}
	A subset $S$ is a global sufficient reason of $f$ if and only if $S$ is a hitting set of $\mathbb{S}$ and is a global contrastive reason of $f$ if and only if $S$ is a hitting set of $\mathbb{C}$.
\end{lemma}

As a consequence of Lemma~\ref{duality_not_minimum}, it directly follows that cardinally minimal local contrastive reasons are with correspondence to MHSs of $\mathbb{S}$, and cardinally minimal local sufficient reasons are with correspondence to MHSs of $\mathbb{C}$.

\textbf{The importance of the MHS duality.} An essential finding when dealing 
with inconsistent sets of clauses lies in a similar MHS duality between Minimal 
Unsatisfiable Sets (MUSes) and Minimal Correction Sets 
(MCSes)~\cite{birnbaum2003consistent, bacchus2015using}. In this context, MCSes 
are MHSs of MUSes, and vice versa~\cite{bailey2005discovery, 
liffiton2008algorithms}. This discovery has played a pivotal role in the 
advancement of algorithms designed for MUSes and MCSes and this result has 
found applications in various contexts~\cite{bacchus2015using, 
liffiton2016fast}. While the majority of this research focuses on propositional 
theories, others focus on Satisfiability Modulo Theories (SMT)~\cite{de2008z3}.

Within the context of explainable AI, previous research has shown similar 
duality principles considering the relationship between \emph{local} 
contrastive and sufficient reasons~\cite{ignatiev2020contrastive} as well as 
the relationship between absolute sufficient reasons and adversarial 
attacks~\cite{ignatiev2019relating}. This relationship was shown to be critical 
in the exact computation of local sufficient reasons for various ML models such 
as decision trees~\cite{izza2022tackling}, tree 
ensembles~\cite{audemard2023computing, audemard2022preferred, izyamajo21, ignatiev2022using, boumazouza2021asteryx}, and neural 
networks~\cite{bassan2023towards, Laagmirhpanikwma21}. In neural networks, 
obtaining exact formal explanations presents substantial computational 
challenges. However, these explanations can be approximated with neural network 
verifiers, which are increasingly utilized for this purpose~\cite{BaAmCoReKa23, 
bassan2023towards, Laagmirhpanikwma21, wu2024verix, huxuma23, fel2023don} and 
for formally verifying other 
properties~\cite{CaKoDaKoKaAmRe22,AmWuBaKa21,AmZeKaSc22,AmFrKaMaRe23,AmMaZeKaSc23}.
 More 
recently,~\cite{BaAmCoReKa23} demonstrated that leveraging the MHS duality may have an even larger effect of efficiency enhancement when computing sufficient reasons in neural 
networks, particularly within the domain of \emph{reactive} systems, a domain in which formal analysis is extensively 
developed~\cite{AmScKa21, 
CoYeAmFaHaKa22,YeAmElHaKaMa22,YeAmElHaKaMa23,AmCoYeMaHaFaKa23, CoAmKaFa24, 
CoAmRoSaKaFo24, MaAmWuDaNeRaMeDuGaShKaBa24}.

\section{The Uniqueness of Global Explanations}
\label{uniquness_proofs}

%\begin{proposition}
%	There exists some $f$ and some $\x\in\mathbb{F}$ such that there are $2^{\floor{\frac{n}{2}}}$ local subset
%	minimal or cardinally minimal sufficient reasons of $\langle f,\x\rangle$.
%\end{proposition}

\begin{proposition}
	There exists some $f$ and some $\x\in\mathbb{F}$ such that there are 
	$\Theta{(\frac{2^n}{\sqrt{n}})}$ local subset
	minimal or cardinally minimal sufficient reasons of $\langle f,\x\rangle$.
\end{proposition}

\emph{Proof.} We construct $f$ as follows:
\begin{equation}
	f(y)=\begin{cases}
		1 \quad if  \ \sum_{i=1}^n y_i\geq \floor{\frac{n}{2}} \\
		0 \quad otherwise
	\end{cases}
\end{equation}
We define the instance $\x:=\mathbf{1}$. Clearly any subset $S$ of size 
$\floor{\frac{n}{2}}$ or larger is a local sufficient reason of $\langle 
f,\x\rangle$ (since fixing the values of $S$ to $\x$ determines that the 
prediction remains: 1). Furthermore, every one of these subsets is 
\emph{minimal} due to the fact that any subset of size $\floor{\frac{n}{2}}-1$ 
or smaller is \emph{not} a sufficient reason of $\langle f,\x\rangle$ (it may 
cause a misclassification to class $0$). Thus, it satisfies that there are 
$\binom{n}{\floor{\frac{n}{2}}}$ subset minimal local sufficient reasons of 
$\langle f,\x\rangle$. From Stirling's approximation, it holds that:
\begin{equation}
    \lim_{n\to \infty} \ \ \frac{2\sqrt{2\pi}}{e^2}\cdot \frac{2^n}{\sqrt{n}} \leq \binom{n}{\floor{\frac{n}{2}}}\leq \frac{e}{\pi}\cdot \frac{2^n}{\sqrt{n}}
\end{equation}

This implies that $\binom{n}{\floor{\frac{n}{2}}}=\Theta{(\frac{2^n}{\sqrt{n}})}$. Given that no 
local 
sufficient reason of size smaller than  $\floor{\frac{n}{2}}$ is present, these 
are, also \emph{cardinally}-minimal sufficient reasons.

\begin{proposition}
	\label{main_core_uniquness}
	If $S_1$ and $S_2$ are two global sufficient reasons of some non-trivial $f$, then
	$S_1\cap S_2 =S \neq \emptyset$, and $S$ is a global sufficient
	reason of $f$.
\end{proposition}

\emph{Proof.} Our proof focuses on \emph{non-trivial} functions, i.e., functions that do not always output $0$ or always output $1$. In other words, there exist some $\x, \mathbf{y}\in\mathbb{F}$ such that $f(\x)=1$ and $f(\mathbf{y})=0$. However, it is important to point out that the general uniqueness property we will prove later (Theorem~\ref{one_unique_unique_appendix_proof}) is independent of whether $f$ is non-trivial.

We begin by proving the following lemma:

\begin{lemma}
	\label{lemma_reverse_sufficient}
	For any $f$ and $\x\in\mathbb{F}$, if $S$ is a sufficient reason of $\langle f,\x\rangle$ then there does not exist any $\mathbf{y}\in\mathbb{F}$ such that $f(\mathbf{y})=\neg f(\x)$ and there exists some $S'\subseteq\overline{S}$ that is a sufficient reason of $\langle f,\mathbf{y}\rangle$. 
\end{lemma}

\emph{Proof.} Given that $S$ is sufficient for $\langle f,\x\rangle$, it follows that:

\begin{equation}
	\forall(\z\in \mathbb{F}).\quad [f(\x_{S};\z_{\Bar{S}})=f(\x)]
\end{equation}

By contradiction, let us assume that there exists some 
$\mathbf{y}\in\mathbb{F}$ for which $f({\mathbf{y}})=\neg f(\x)$ and there 
exists some $S'\subseteq \overline{S}$ that is a sufficient reason of $\langle 
f,\mathbf{y}\rangle$. Since $S'\subseteq \overline{S}$ is a sufficient reason 
for $\langle f,\mathbf{y}\rangle$, this also implies that $\overline{S}$ is 
sufficient for $\langle f,\mathbf{y}\rangle$. In other words, the following 
condition holds:

\begin{equation}
	\label{equation_for_appendix_proof_lemma}
	\exists (\mathbf{y}\in\mathbb{F}), \ \forall(\z\in \mathbb{F}).\quad [f(\mathbf{y}_{\Bar{S}};\z_{S})=f(\mathbf{y})\neq f(\x)]
\end{equation}

Given that Equation~\ref{equation_for_appendix_proof_lemma} is valid for any $\z \in \mathbb{F}$, it is, consequently, applicable specifically to $\x\in\mathbb{F}$. In other words:
\begin{equation}
	\exists (\mathbf{y}\in\mathbb{F})\quad [f(\mathbf{y}_{\Bar{S}};\x_{S})=f(\mathbf{y})\neq f(\x)]
\end{equation}

This is inconsistent with the assertion that $S$ is sufficient for $\langle f,\x\rangle$.

\begin{lemma}
	\label{non_trivial_2}
	For a non-trivial function $f$, if $S$ is a sufficient reason of $\langle f,\x\rangle$ then any $S'\subseteq\overline{S}$ is not a global sufficient reason of $f$. 
\end{lemma}

\emph{Proof.} 
Given that $S$ serves as a sufficient reason for $\langle f, \mathbf{x} \rangle$, it follows from Lemma~\ref{lemma_reverse_sufficient} that there does not exist any $\mathbf{y} \in \mathbb{F}$ for which $f(\mathbf{y}) = \neg f(\mathbf{x})$ and $\overline{S}$ is sufficient for $\langle f, \mathbf{y} \rangle$. Consequently, if there indeed exists some $\mathbf{y} \in \mathbb{F}$ for which $\overline{S}$ serves as a sufficient reason for $\langle f, \mathbf{y} \rangle$, it necessarily follows that $f(\mathbf{x}) = f(\mathbf{y})$.

Let us, by contradiction, assume the existence of some 
$S'\subseteq\overline{S}$ that serves as a global sufficient reason of $f$. 
This implication further entails that $\overline{S}$ is also a global 
sufficient reason for $f$. Consequently, from the definition of global sufficient reasons, $\overline{S}$ is also a local 
sufficient reason for $\langle f,\mathbf{y}\rangle$ for any 
$\mathbf{y}\in\mathbb{F}$. Given the property highlighted earlier, it holds 
that for any $\mathbf{y}\in\mathbb{F}$, we have $f(\mathbf{y})=f(\x)$, which stands in 
contradiction to the premise that $f$ is non-trivial.

We are now in a position to prove the first part of proposition~\ref{main_core_uniquness}:

\begin{lemma}
	\label{one_before_last_first_part_long_proposition}
	If $S_1$ and $S_2$ are two global sufficient reasons of some non-trivial $f$, then
	$S_1\cap S_2 \neq \emptyset$.
\end{lemma}

\emph{Proof.} Let us assume, to the contrary, that $S_1\cap S_2=\emptyset$. Hence, it follows that $S_1\subseteq \overline{S_2}$. Given that $S_2$ is a global sufficient reason for $f$, it naturally follows that it is also a local sufficient reason for some $\langle f,\x\rangle$. Since $S_2$ is a local sufficient reason for some $\langle f,\x\rangle$, Lemma~\ref{non_trivial_2} determines that there does not exist any $S'\subseteq \overline{S_2}$ that can be a global sufficient reason for $f$. This is in direct contradiction with our earlier inference that $S_1\subseteq\overline{S_2}$ is a global sufficient reason for $f$.

We now can proceed to prove the second part of proposition~\ref{main_core_uniquness}:

\begin{lemma}
	\label{the_main_shit}
	If $S_1$ and $S_2$ are global sufficient reasons of some non-trivial $f$ %and $S_1\not\subseteq S_2$, $S_2\not\subseteq S_1$, 
	,then $S=S_1\cap S_2$ is a global sufficient reason of $f$.
\end{lemma}

\emph{Proof.} First, from Lemma~\ref{one_before_last_first_part_long_proposition}, it holds that $S\neq \emptyset$. In instances where either $S_1\subseteq S_2$ or $S_2\subseteq S_1$, the claim is straightforwardly true. Therefore, our remaining task is to prove the claim for a non-empty set $S$ with the conditions $S\subsetneq S_1$ and $S\subsetneq S_2$.

Consider an arbitrary vector $\x\in\mathbb{F}$. Our aim is to prove that $S$ is a local sufficient reason with respect to $\langle f,\x\rangle$. Should this hold true for an arbitrary $\x$, it follows that $S$ constitutes a global sufficient reason of $f$. 

Given that $S_1$ and $S_2$ are global sufficient reasons, it holds that:

%denote arbitrary value assignments as follows: $V_1$ are the corresponding assignments to $S$, $V_2$ are the assignments to $S_1\setminus S$ and $V_3$ are the assignments to $S_2\setminus S$. Since $S_1$ is a global sufficient reason it holds that fixing the values of $S$ to $V_1$ and of $S_1\setminus S$ to $V_2$ determines that the prediction remains the same, regardless of the features of $\bar{S_1}$. 

\begin{equation}
	\label{explanation_s1}
	\forall(\z\in \mathbb{F}).\quad [f(\x_{S_1};\z_{\Bar{S_1}})=f(\x)=f(\x_{S_2};\z_{\Bar{S_2}})]
\end{equation}

To demonstrate that $S$ is a local sufficient reason for $\langle f,\x\rangle$, 
let us assume, by contradiction, that it is not. Therefore, it satisfies that:

\begin{equation}
	\begin{aligned}
	\label{equation_in_process_for_reference_text}
	\exists(\z\in \mathbb{F}).\quad [f(\x_{S};\z_{\Bar{S}})\neq f(\x)]  \iff \\
	\exists(\z\in \mathbb{F}).\quad [f(\x_{S};\z_{S_2\setminus S};\z_{\Bar{S_2}})\neq f(\x)]
	\end{aligned}
\end{equation}

Recall that $S_2$ is a global sufficient reason of $f$. Thus, assigning the features of $S$ to the corresponding values in $\x$ determines the prediction. This also implies that fixing the features of $S$ to the corresponding values $\x$ and assigning those of $S_2\setminus S$ to the specific values of $\z$ from equation~\ref{equation_in_process_for_reference_text}, determines that the prediction remains the same (which in this case is \emph{not} the value $f(\x)$). Formally put:

\begin{equation}
	\label{equation_reference_for_proof}
	\forall(\z'\in \mathbb{F}).\quad [f(\x_{S};\z_{S_2\setminus S};\z'_{\Bar{S_2}})=f(\x_{S};\z_{S_2\setminus S};\z_{\Bar{S_2}})\neq f(\x)]
\end{equation}

Let us define the set $S'=\{1,\ldots,n\}\setminus\{S_1\cup S_2\}$. We now can equivalently express equation~\ref{equation_reference_for_proof} as:

\begin{equation}
	\label{one_before_last}
	\forall(\z'\in \mathbb{F}).\quad [f(\x_{S};\z_{S_2\setminus S};\z'_{S_1\setminus S};\z'_{S'})\neq f(\x)]
\end{equation}

But we know that $S_1$ is also a global sufficient reason and hence fixing the values of $S_1$ to $\x$ determines that the prediction is $f(\x)$. Particularly, fixing the values of $S_1$ to $\x$ and the values of $S_2\setminus S$ to $\z$ (from equation~\ref{equation_in_process_for_reference_text}) still determines that the prediction is always $f(\x)$.

\begin{equation}
	\label{explanation_s1}
	\forall(\z'\in \mathbb{F}).\quad [f(\x_{S_1};\z_{S_2\setminus S};\z'_{S'\cup S})=f(\x)]
\end{equation}

Since the preceding statement remains valid for any partial assignment of the features in $S'\cup S$, we can consider a specific assignment where the features in $S$ are set to their respective values in $\x$; thus, it is established that:

\begin{equation}
	\begin{aligned}
	\label{explanation_s1}
	\forall (\z'\in \mathbb{F}).\quad [f(\x_S;\x_{S_1};\z_{S_2\setminus S};\z'_{S'})=f(\x)]
	\end{aligned}
\end{equation}

This particularly implies that:

\begin{equation}
	\begin{aligned}
        \exists (\z'\in \mathbb{F}).\quad [f(\x_{S};\z_{S_2\setminus S};\x_{S_1\setminus S};\z'_{S'})=f(\x)]
	\end{aligned}
\end{equation}

which contradicts Equation~\ref{one_before_last}.

\begin{theorem}
	\label{one_unique_unique_appendix_proof}
	There exists one unique subset-minimal global sufficient reason of $f$.
\end{theorem}

%We assume that there exists an input for which the classification is either $0$ or $1$ (it does not hold that all inputs are classified to only one of them).

\emph{Proof.} First, for the scenario in which $f$ is trivial (always outputs 
$1$ or always outputs $0$) it holds that any subset $S$ is a global sufficient 
reason. Therefore, $S=\emptyset$ is a unique subset-minimal global sufficient 
reason. Let us now consider a non-trivial function $f$. Let us assume, by 
contradiction, that two distinct subset minimal global sufficient 
reasons of $f$ exist: $S_1\neq S_2$. Since $S_1$ and $S_2$ are subset minimal, 
it clearly holds that $S_1\not\subseteq S_2$ and $S_2\not\subseteq S_1$. 
Moreover, from Proposition~\ref{two_opposite_globals_dont_exist} it can be 
asserted that $S_1$ and $S_2$ are not disjoint, i.e., $S_1\cap S_2\neq 
\emptyset$. Now, we can use Proposition~\ref{the_main_shit}, and conclude that 
$S=S_1\cap S_2$ is also a global sufficient reason of $f$. This clearly 
contradicts the subset minimality of $S_1$ and $S_2$.

\begin{proposition}
	\label{any_ordering_converges_appendix}
	For any possible ordering of features in
	line~\ref{lst:line:orderinglineglobal} of
	Algorithm~\ref{alg:subset-minimal-global},
	Algorithm~\ref{alg:subset-minimal-global} converges to the same
	global sufficient reason.
\end{proposition}

Since Algorithm~\ref{alg:subset-minimal-global} converges to a subset-minimal global sufficient reason, and there is only one unique subset-minimal global sufficient reason (Theorem~\ref{one_unique_unique_appendix_proof}), then iterating over any ordering of features in
line~\ref{lst:line:orderinglineglobal} of Algorithm~~\ref{alg:subset-minimal-global} will converge to the same subset. 
We note that the convergence of Algorithm~\ref{alg:subset-minimal-global} to a subset-minimal sufficient reason stems from the hereditary property of sufficient reasons, applicable to both local and global contexts. Specifically, if $S \subseteq \{1, \ldots, n\}$ is a (local/global) sufficient reason, then any subset $S' \subseteq \{1, \ldots, n\}$ that contains $S$ (i.e., $S \subseteq S'$) will also be a (local/global) sufficient reason.

\begin{proposition}
	\label{last_proposition_uniquenss_appendix}
	Let $S$ be the unique subset minimal global sufficient reason of
	$f$. For every $i$, $i\in S$ if and only if $i$ is locally necessary
	for some $\langle f,\x\rangle$, and $i\in \overline{S}$ if and only
	if $i$ is globally redundant for $f$.
\end{proposition}

We begin by proving the first part of the claim:

\begin{lemma}
	\label{first_lemma_last_part_uniquness}
	$S$ is a unique subset-minimal global sufficient reason of $f$ if and only 
	if for every feature $i\in S$ it holds that $i$ is locally necessary for 
	some $\langle f,\x\rangle$.
\end{lemma}

\emph{Proof.} For the first direction, assume $i$ is necessary for some 
$\langle f,\x\rangle$. Then, from Theorem~\ref{necessitiy_contrastive_iff}, it 
holds that $\{i\}$ is contrastive for some $\langle f,\x\rangle$. Furthermore, 
the first duality theorem (Theorem~\ref{duality_one_appendix}), implies that 
each local contrastive reason intersects each global sufficient reason. Hence, 
we conclude that $i\in S$, for every global minimal sufficient reason S.

For the second direction, suppose that $S$ is a unique subset-minimal global 
sufficient reason of $f$. Let there be some $i\in S$. Since $S$ is 
\emph{unique}, then $\{1,\ldots,n\}\setminus \{i\}$ is necessarily \emph{not} a 
global sufficient reason. If this was so, then there would exist some subset 
$S'\subseteq \{1,\ldots,n\}\setminus \{i\}$ that is a subset-minimal global 
sufficient reason, contradicting the uniqueness of $S$.

Since $\{1,\ldots n\}\setminus \{i\}$ is not a global sufficient reason, there exist some $\x',\z'\in \mathbb{F}$ such that:

\begin{equation}
	f(\x'_{\{1,\ldots n\}\setminus \{i\}};\z'_{\{i\}})\neq f(\x')
\end{equation}

Thus, $\{i\}$ serves as a contrastive reason for $\langle f,\x'\rangle$ and from Theorem~\ref{necessitiy_contrastive_iff} we can infer that $i$ is necessary with respect to $\langle f,\x'\rangle$.

For the second part of the claim, we prove the following Lemma:

\begin{lemma}
	Let $S$ be the unique subset-minimal global sufficient reason of $f$. Then $i$ is globally redundant if and only if $i\in \overline{S}$.
\end{lemma}

\emph{Proof.} For the first direction, let us assume that $i$ is globally 
redundant and assume, by contradiction, that $i\in S$. Given that $i$ is 
globally redundant for $f$ then it holds that for any $\x\in\mathbb{F}$: 
$\mySuff(f,\x,S)=1\to \mySuff(f,\x,S\setminus \{i\})=1$. Hence, $S\setminus 
\{i\}$ is also a global sufficient reason of $S$, contradicting the 
subset-minimality of $S$.

For the second direction, assume that $i\in \overline{S}$. From Lemma~\ref{first_lemma_last_part_uniquness}, it holds that $i$ is not locally necessary for any $\langle f,\x\rangle$. In other words, there does not exist any $\mathbf x\in\mathbb{F}$ for which $\mySuff(f,\x,S)=1\to \mySuff(f,\x,S\setminus \{i\})=0$. This implies that for any $\x\in\mathbb{F}$ it satisfies that $\mySuff(f,\x,S)=1\to \mySuff(f,\x,S\setminus \{i\})=1$, i.e., that $i$ is globally redundant with respect to $f$.

\begin{proposition}
	\label{split_necessary_redundant_appendix}
	Any feature $i$ is either locally necessary for some $\langle f,\x\rangle$ or globally redundant for $f$.
\end{proposition}

Building upon Proposition~\ref{last_proposition_uniquenss_appendix}, we can 
discern that the unique subset minimal sufficient reason $S$ of $f$ categorizes 
all features into two distinct categories: those that are locally necessary for 
some $\langle f,\x\rangle$ and those that are globally redundant for $f$.

\section{Proof of proposition~\ref{proposition_csr_appendix}}
\label{csr_appendix}

\begin{proposition}
	\label{proposition_csr_appendix}
For Perceptrons, solving G-CSR and G-MSR is coNP-Complete, while solving CSR and MSR can be done in polynomial time.
\end{proposition}

        As mentioned in Table~\ref{table:results:VerificationResults}, the complexity of \emph{CSR} and \emph{MSR} for Perceptrons are drawn from the work of~\cite{BaMoPeSu20} (similar proofs appear in~\cite{marques2020explaining}). We now move to prove all other complexity classes:

\begin{lemma}
	\label{csr_lemma_perceptrons_appendix}
	G-CSR is coNP-Complete for Perceptrons.
\end{lemma}
\emph{Proof.} Membership is straightforward since we can simply guess some $\x,\z\in\mathbb{F}$ and validate whether it satisfies that $f(\x_{S};\z_{\Bar{S}}) \neq f(\x)$. If so, $\langle f, S\rangle\not\in$ \emph{G-CSR}.

We now will proceed to prove that \emph{G-CSR} is also coNP-hard, We first 
briefly describe how the problem of (local) \emph{CSR} can be solved in 
polynomial time for perceptrons, as proven by~\cite{BaMoPeSu20} (a similar proof for a general linear classifier appears in~\cite{marques2020explaining}). This will give 
better intuition for the hardness reduction in the global setting. Given some 
$\langle f,\x,S\rangle$, recall that a Perceptron $f$ is defined as $f=\langle 
\w,b\rangle$, where $\w$ is the weight vector and $b$ is the 
bias term. Therefore, it is possible to obtain the exact value of $\sum_{i\in 
S}\x_i\cdot \w_i$. 

Then, for the remaining features in $\overline{S}$, one can linearly determine the 
$\y$ assignments corresponding to the \emph{maximal} and \emph{minimal} values 
of $\sum_{i\in \Bar{S}}\y_i\cdot \w_i$. The maximal value is obtained by setting 
$\y_i:=1$ when $\w_i\geq 0$ and $\y_i:=0$ when $\w_i\leq 0$. 
The minimal value is obtained respectively (setting $\y_i:=1$ when 
$\w_i< 0$ and $\y_i:=0$ when $\w_i\geq0$). We are now 
able to compute the full range of potential values 
that may be realized by assigning the values of $S$ to $\x$. It is hence 
straightforward that $S$ is a (local) sufficient reason for $\langle 
f,\x\rangle$ if and only if this entire range is always positive or always 
negative. This can be determined by checking whether both the minimal possible 
value and maximal possible value are both positive or negative which is 
equivalent to checking whether the maximal possible value is non-negative or 
that the minimal possible value is positive. Formally put:

\begin{equation}
	\begin{aligned}
		\label{perceptrons_csr_local_equation_to_cite}
		\sum_{i\in S}\x_i\cdot \w_i+max\{\sum_{i\in \Bar{S}}\y_i\cdot \w_i+b\ | \ \y\in \mathbb{F}\}\leq0 \ \ \vee \\
		\sum_{i\in S}\x_i\cdot \w_i+min\{\sum_{i\in \Bar{S}}\y_i\cdot \w_i+b\ | \ \y\in \mathbb{F}\}>0 
	\end{aligned}
\end{equation}

This can clearly be determined in linear time using the computation method 
outlined above. Note that we require a strict inequation on the second term 
since we assumed w.l.o.g. that a zero weighted term is classified as $0$ (the 
negative weighted class) and not $1$ (the positive weighted class).

Now, for the global setting, we notice that $max\{\sum_{i\in \Bar{S}}\y_i\cdot \w_i+b\ | \ \y\in \mathbb{F}\}$ and $min\{\sum_{i\in \Bar{S}}\y_i\cdot \w_i+b\ | \ \y\in \mathbb{F}\}$ can still be computed in the same manner as above. However, one must verify that equation~\ref{perceptrons_csr_local_equation_to_cite} is satisfied for \emph{every} possible value $\x$. This, in turn, carries implications for the associated complexity. We show, indeed, that \emph{G-CSR} for perceptrons is coNP-hard.

We reduce \emph{G-CSR} for Perceptrons from $\overline{SSP}$, known to be coNP-Complete. \emph{SSP} (subset-sum-problem) is a classic NP-Complete problem which is defined as follows:

\noindent\fbox{%
	\parbox{\textwidth}{%
		\mysubsection{\textbf{SSP} (Subset Sum Problem)}:
		
		\textbf{Input}: $\langle (\z_1,\z_2,\ldots,\z_n), T\rangle$, where $(\z_1,\z_2,\ldots,\z_n)$ is a set of \emph{positive integers} and $T$, is the target integer.
		
		\textbf{Output}: \yes{}, if there exists a subset $S'\subseteq \{1,2,\ldots,n\}$ such that $\sum_{i\in S'}\z_i=T$, and \no{} otherwise.
	}%
}

For the case of $\overline{SSP}$, the language decides whether there does not 
exist a subset of features $S'\subseteq (1,2,\ldots,n)$ for which $\sum_{i\in 
S'}\z_i=T$, i.e., for \emph{all} subsets it holds that $\sum_{i\in S'}\z_i\neq T$.

We reduce \emph{G-CSR} for Perceptrons from $\overline{SSP}$. Given some 
$\langle (\z_1,\z_2,\ldots,\z_n),T\rangle$ we construct a Perceptron $f:=\langle 
\w,b\rangle$ such that $\w:=(\z_1,\z_2,\ldots,\z_n)\cdot 
(\mathbf{w}_{n+1})$ ($\w$ is of size $n+1$), where 
$\w_{n+1}:=\frac{1}{2}$, and $b:=-(T+\frac{1}{4})$. The reduction 
computes $\langle f,S:=\{1,\ldots,n\}\rangle$. 

Clearly, it holds that: 

\begin{equation}
	\begin{aligned}
 \label{first_first_first_equation}
		max\{\sum_{i\in \Bar{S}}\y_i\cdot \w_i\ | \ \y\in \mathbb{F}\} =max\{\sum_{i\in \{1,\ldots,n+1\}\setminus \{1,\ldots,n\} }\y_i\cdot \w_i\ | \ \y\in \mathbb{F}\}=\\
  max\{\sum_{i=n+1}\y_i\cdot \w_i\ | \ \y\in \mathbb{F}\}=
  max\{\frac{1}{2},0\}=  \frac{1}2
	\end{aligned}
\end{equation}

and that:

\begin{equation}
	\begin{aligned}
 \label{the_equation_before_to_cite}
		min\{\sum_{i\in \Bar{S}}\y_i\cdot \mathbf{w}_i\ | \ \y\in \mathbb{F}\} =min\{\sum_{i\in \{1,\ldots,n+1\}\setminus \{1,\ldots,n\} }\y_i\cdot \w_i\ | \ \y\in \mathbb{F}\}=\\
  min\{\sum_{i=n+1}\y_i\cdot \w_i\ | \ \y\in \mathbb{F}\}= min\{\frac{1}{2},0\}= 0
	\end{aligned}
\end{equation}

If $\langle (\z_1,\z_2,\ldots,\z_n),T\rangle\in\overline{SSP}$, there does not exist a subset $S'\subseteq S = \{1,2,\ldots,n\}$ for which $\sum_{i\in S'}\z_i=T$, put differently --- for any subset $S'\subseteq S = \{1,2,\ldots,n\}$ it holds that $\sum_{i\in S'}\z_i>T$ or $\sum_{i\in S'}\z_i<T$. But since the values in $(\z_1,\z_2,\ldots,\z_n)$ are \emph{positive integers} then it also holds that for any subset $S'$ the following condition is met:

\begin{equation}
	\begin{aligned}
		\label{perceptrons_csr_local_equation}
		\forall S'\subseteq S \ \ \  [ \ [\sum_{i\in S'}\z_i>T+\frac{1}{4}] \ \ \vee [\sum_{i\in S'}\z_i<T-\frac{1}{4}] \ ] \iff \\
  		\forall S'\subseteq S \ \ \ [ \ [\sum_{i\in S'}\w_i>T+\frac{1}{4}] \ \ \vee [\sum_{i\in S'}\w_i<T-\frac{1}{4}] \ ] \iff \\
      	\forall S'\subseteq S \ \ \ [ \ [\sum_{i\in S'}\w_i\cdot \mathbf{1}_i + \sum_{i\in S\setminus S'}\w_i\cdot \mathbf{0}_i >T+\frac{1}{4}] \ \ \vee [\sum_{i\in S'}\w_i\cdot \mathbf{1}_i + \sum_{i\in S\setminus S'}\w_i\cdot \mathbf{0}_i<T-\frac{1}{4}] \ ] \iff \\
		%\forall \x\in\mathbb{F} \ \ \ [[\sum_{i\in S'}\x_i\cdot w_i> T+\frac{1}{4}] \ \ \vee \ \
		%[\sum_{i\in S'}\x_i\cdot w_i< T-\frac{1}{4}]] \iff \\
		\forall \x\in\{0,1\}^n \ \ \ [ \ [\sum_{i\in S}\x_i\cdot \w_i> T+\frac{1}{4}] \ \ \vee \ \
		[\sum_{i\in S}\x_i\cdot \w_i< T-\frac{1}{4}]\ ] \iff \\
  		\forall \x\in\{0,1\}^n \ \ \ [ \ [\sum_{i\in S}\x_i\cdot \w_i+b> 0] \ \ \vee \ \
		[\sum_{i\in S}\x_i\cdot \w_i+b<-\frac{1}{2}]\ ]%\iff \\
    	%	\forall \x\in\{0,1\}^{n+1} \ \ \ [ \ [\sum_{i\in S}\x_i\cdot w_i+\x_{n+1}\cdot w_{n+1}+b> 0] \ \ \vee \ \
	%	[\sum_{i\in S}\x_i\cdot w_i+\x_{n+1}\cdot w_{n+1}+b<0]\ %] \iff \\
        %\forall \x\in\{0,1\}^{n+1}, \forall \z\in\{0,1\}^{n+1} \ \ \ [ \ [f(\x_S;\z_{\Bar{S}})> 0] \ \ \vee \ \
	%	[f(\x_S;\z_{\Bar{S}})<0]\ ]
	\end{aligned}
\end{equation}

Since we know from equations~\ref{first_first_first_equation} and~\ref{the_equation_before_to_cite} that: 

\begin{equation}
\begin{aligned}
    [ \  min\{\sum_{i\in \Bar{S}}\y_i\cdot \mathbf{w}_i\ | \ \y\in \mathbb{F}\}= 0] \ \ \ \wedge 
    [ \ max\{\sum_{i\in \Bar{S}}\y_i\cdot \w_i\ | \ \y\in \mathbb{F}\} =\frac{1}{2} \ ]
    \end{aligned}
\end{equation}

This, combined with the result from equation~\ref{perceptrons_csr_local_equation}, implies that:

\begin{equation}
\begin{aligned}
  		\forall \x\in\{0,1\}^{n} \ \ \ [ \ [\sum_{i\in S}\x_i\cdot \w_i+b> 0] \ \ \vee \ \
		[\sum_{i\in S}\x_i\cdot \w_i+b<-\frac{1}{2}]\ ] \iff \\
        \forall \x\in\{0,1\}^{n+1} \ \ \ [ \ [\sum_{i\in S}\x_i\cdot \w_i+\x_{n+1}\cdot \w_{n+1}+b> 0] \ \ \vee \ \
		[\sum_{i\in S}\x_i\cdot \w_i+\x_{n+1}\cdot \w_{n+1}+b<0]\ ] \iff \\
          \forall \x,\y\in\{0,1\}^{n+1} \ \ \ [ \ [\sum_{i\in S}\x_i\cdot \w_i+\sum_{i\in \overline{S}}\y_i\cdot \w_i+b> 0] \ \ \vee \ \
		[\sum_{i\in S}\x_i\cdot w_i+\sum_{i\in \overline{S}}\y_i\cdot \w_i+b<0]\ ] \iff \\
          \forall \x,\y\in\{0,1\}^{n+1} \ \ \ [ \ [f(\x_S;\y_{\Bar{S}})=f(\x_S;\neg\y_{\Bar{S}})= 0] \ \ \vee \ \
		[f(\x_S;\y_{\Bar{S}})=f(\x_S;\neg\y_{\Bar{S}})=1]\ ]
\end{aligned}
\end{equation}

This implies that for any $\x\in\mathbb{F}$, fixing the values of $S=\{1,\ldots,n\}$ always maintains either a positive value (classified to $1$) or a negative value (classified to $0$) for $f$ over any possible $\y$, thus implying that $S$ is a global sufficient reason of $f$ and that $\langle f,S\rangle \in$ \emph{G-CSR}.

If $\langle (\z_1,\z_2,\ldots,\z_n),T\rangle\not\in \overline{SSP}$, then there exists a subset $S'\subseteq S=\{1,2,\ldots,n\}$ for which $\sum_{s_i\in S'}\z_i=T$, implying that:

\begin{equation}
	\begin{aligned}
		\label{perceptrons_csr_local}
		\exists S'\subseteq S \ \ \ [T-\frac{1}{4}]<\sum_{i\in S'}\z_i< [T+\frac{1}{4}] \iff 
  		\exists S'\subseteq S \ \ \ [T-\frac{1}{4}]<\sum_{i\in S'}\w_i< [T+\frac{1}{4}] \iff \\
        \exists S'\subseteq S \ \ \ [T-\frac{1}{4}]<\sum_{i\in S'}\w_i\cdot \mathbf{1}_i+\sum_{i\in S\setminus S'}\w_i\cdot \mathbf{0}_i< [T+\frac{1}{4}] \iff \\
          	\exists \x\in\{0,1\}^n \ \ \ [T-\frac{1}{4}]<\sum_{i\in S}\w_i\cdot \x_i< [T+\frac{1}{4}]\iff          \exists \x\in\{0,1\}^n \ \ \ (-\frac{1}{2})<\sum_{i\in S}\w_i\cdot \x_i+b< 0
	\end{aligned}
\end{equation}

From here it holds that:

\begin{equation}
\begin{aligned}
    [ \ \exists \x\in\{0,1\}^{n} \ \ \ 0<\sum_{i\in S}\w_i\cdot \x_i + b+ \w_{n+1}\cdot 1 < (\frac{1}{2}) \ ] \ \ \wedge  \\    [ \    \exists \x\in\{0,1\}^{n} \ \ \ (-\frac{1}{2})<\sum_{i\in S}\w_i\cdot \x_i +b+ \w_{n+1}\cdot 0 < 0 \ ]
    \end{aligned}
\end{equation}

This implies that:

\begin{equation}
\begin{aligned}
     \ \exists \x,\y,\y'\in\{0,1\}^{n+1} \ \ [ \ 0<\sum_{i\in S}\w_i\cdot \x_i +\w_{n+1}\cdot \y_{n+1}+ b < (\frac{1}{2}) \ ] \ \wedge \\ [\ -(\frac{1}{2})<\sum_{i\in S}\w_i\cdot \x_i +\w_{n+1}\cdot \y'_{n+1}+  b < 0 ]
    \end{aligned}
\end{equation}

This is equivalent to saying there exist features $\x,\y,\y'\in\{0,1\}^{n+1}$ such that the output of the perceptron $f$ for the input $(\x_S;\y_{\Bar{S}})$ always lies in the range from $0$ to $\frac{1}{2}$ (thereby being positive), and the output for the input $(\x_S;\y'_{\Bar{S}})$ ranges from $-(\frac{1}{2})$ to $0$ (thereby being negative). Hence:

\begin{equation}
         \ \exists \x,\y,\y'\in\{0,1\}^{n+1} \ \ [ \  f(\x_{S};\y_{\Bar{S}})=1 ] \ \wedge [\ f(\x_{S};\y'_{\Bar{S}}) = 0 ]
\end{equation}

Thus, there exists some $\x$ for which $S$ is not a sufficient reason with respect to $\langle f,\x\rangle$, indicating that $S$ is not a global sufficient reason for $f$ and $\langle f,S\rangle \not\in$ \emph{G-CSR}. This concludes the reduction.

Hardness results for Perceptrons, clearly indicate coNP-hardness for MLPs.

\begin{lemma}
	G-MSR is coNP-Complete for Perceptrons.
\end{lemma}

\emph{Proof.}
\textbf{Membership.} 
Membership is derived from the fact that one can guess some $\x^1,\ldots 
\x^n\in \mathbb{F}$ and $\z^1,\ldots,\z^n\in\mathbb{F}$. We then can validate 
for every feature $i\in (1,\ldots,n)$ whether: $f(\x^i_{\{1,\ldots,n\}\setminus 
\{i\}};\z^i_{\{i\}})\neq f(\x^i)$. This will imply that $\{i\}$ is contrastive 
with respect to $\langle f,\x^i\rangle$ and from 
Theorem~\ref{necessetiy_contrastive_theorem}, $i$ is necessary with respect to 
$\langle f,\x^i\rangle$. Now, from 
Proposition~\ref{last_proposition_uniquenss_appendix} it holds that $i$ is 
contained in the unique global subset minimal sufficient reason of $f$ if and 
only if $i$ is necessary with respect to some $\langle f,\x\rangle$. Therefore, 
it is possible to validate whether $\langle f,k\rangle \not\in$ \emph{G-MSR} 
using a certificate that checks whether the number of features that satisfy: 
$f(\x^i_{\{1,\ldots,n\}\setminus \{i\}};\z^i_{\{i\}})\neq f(\x^i)$ is larger 
than $k$.

\textbf{Hardness.} For hardness, we perform a similar reduction to the one 
performed for \emph{G-CSR} for Perceptrons and reduce \emph{G-MSR} for 
Perceptrons from $\overline{SSP}$. Given some $\langle 
(\z_1,\z_2,\ldots,\z_n),T\rangle$, construct a Perceptron $f:=\langle 
\mathbf{w},b\rangle$ where we define $\mathbf{w}:=(\z_1,\z_2,\ldots,\z_n)\cdot 
(\mathbf{w}_{n+1})$ (i.e., $\mathbf{w}$ is of size $n+1$), where 
$\mathbf{w}_{n+1}:=\frac{1}{2}$, and $b:=-(T+\frac{1}{4})$. The reduction 
computes $\langle f,k:=n\rangle$. We observe that this reduction is precisely the same as the one presented in Lemma~\ref{csr_lemma_perceptrons_appendix}, with the only difference being that we define $k:=n$ rather than $S:=\{1,\ldots,n\}$. Therefore, we will base some of our assertions on the validity of the reduction from Lemma~\ref{csr_lemma_perceptrons_appendix}.

Consider that $\langle (\z_1,\z_2,\ldots,\z_n),T\rangle\in \overline{SSP}$. 
Drawing upon the correctness of the reduction in Lemma~\ref{csr_lemma_perceptrons_appendix}, we can infer that 
$S=\{1,2,\ldots,n\}$ constitutes a global sufficient reason of $f$. To put it 
differently, a subset exists --- trivially of size $k=n$ in this instance 
--- that serves as a global sufficient reason of $f$. Consequently, $\langle 
f,k\rangle\in$ \emph{G-MSR} for Perceptrons. 

Assume $\langle (\z_1,\z_2,\ldots,\z_n),T\rangle\not\in \overline{SSP}$. We need 
to prove that there does not exist any global sufficient reason of $f$ of size 
$k$ or less. Since any subset containing a sufficient reason is a sufficient 
reason, it is enough to show that there does not exist any global sufficient 
reason of \emph{exactly} size $k$. From the correctness of the reduction presented in
Lemma~\ref{csr_lemma_perceptrons_appendix} we indeed already know that $\{1,\ldots,n\}$, which corresponds to the assignments of
$(\z_1,\z_2,\ldots,\z_n)$ is \emph{not} a sufficient reason in this case. However, 
we still need to prove that there does not exist any \emph{other} sufficient 
reason of size $k$.

Let $j\neq n+1$ be some feature and let $S:=\{1,2,\ldots,n+1\}\setminus \{j\}$ 
be some subset of features. We will now prove that $S$ is not a global sufficient reason 
for $f$. 
First, since any $\z_j$ in $(\z_1,\z_2,\ldots,\z_n)$ is a positive integer, and since 
$\w_{n+1}=\frac{1}{2}$ is also positive, then it holds that:

\begin{equation}
	\begin{aligned}
		max\{\sum_{i\in \Bar{S}}\y_i\cdot \w_i\ | \ \y\in \mathbb{F}\} = 
		max\{\z_j,0\}= \z_j \ \ \wedge \\
		min\{\sum_{i\in \Bar{S}}\y_i\cdot \w_i\ | \ \y\in \mathbb{F}\} = 
		min\{\z_j,0\}= 0
	\end{aligned}
\end{equation}

This implies that that $S$ is a global sufficient reason of $f$ iff: %for any 
%$\x\in\mathbb{F}$ it holds that:

\begin{equation}
	\label{important_equation_proof}
	\begin{aligned}
		\forall \x\in\{0,1\}^{n+1} \ \ \ [\sum_{i\in S}\x_i\cdot \w_i > T] \ \ \vee \  \
		[\sum_{i\in S}\x_i\cdot \w_i \leq T-\z_j]
	\end{aligned}
\end{equation}

Within Lemma~\ref{csr_lemma_perceptrons_appendix} we have already determined 
that if $\langle (\z_1,\z_2,\ldots,\z_n),T\rangle\not\in \overline{SSP}$, then the following holds: 
%there must exist a value $\x\in\mathbb{F}$ such that:

\begin{equation}
	\label{equation_use_this_now}
	\begin{aligned}
	\exists \x\in\{0,1\}^n \ \ \	\sum_{i\in \{1,\ldots,n\}}\x_i\cdot \w_i = T \iff \\
	\exists \x\in\{0,1\}^n \ \ \	\sum_{i\in \{1,\ldots,n\}\setminus \{j\}}\x_i\cdot \w_i = T-\z_j \iff \\
 	\exists \x\in\{0,1\}^n \ \ \	\sum_{i\in \{1,\ldots,n\}\setminus \{j\}}\x_i\cdot \w_i +\w_{n+1} = T-\z_j+\frac{1}{2} \iff \\
	\exists \x\in\{0,1\}^{n+1} \ \ \	\sum_{i\in \{1,\ldots,n,n+1\}\setminus \{j\}}\x_i\cdot \w_i = 
		T-\z_j+\frac{1}{2}
	\end{aligned}
\end{equation}

Now, since $T$ and $\z_j$ are positive \emph{integers}, then from 
equation~\ref{equation_use_this_now} it holds that: %there exists some instance 
%$\x\in\mathbb{F}$ such that:

\begin{equation}
	\begin{aligned}
 	\exists \x\in\{0,1\}^{n+1} \ \ \	\sum_{i\in \{1,\ldots,n,n+1\}\setminus \{j\}}\x_i\cdot w_i = 
		T-\z_j+\frac{1}{2} \ \Rightarrow \\
	\exists \x\in\{0,1\}^{n+1} \ \ \	T-\z_j<\sum_{i\in \{1,\ldots,n,n+1\}\setminus \{j\}}\x_i\cdot \w_i < T
	\end{aligned}
\end{equation}

From equation~\ref{important_equation_proof}, this implies that 
$\{1,\ldots,n,n+1\}\setminus \{j\}$ is \emph{not} a global sufficient reason of 
$f$. Since there does not exist any $j$ for which $\{1,\ldots,n,n+1\}\setminus 
\{j\}$ is a global sufficient reason of $f$ and since we have already 
determined that $\{1,\ldots,n\}$ is not a global sufficient reason of $f$, we 
are left with that there does not exist any global sufficient reason of size 
$k$, concluding the reduction.

\section{Proof of Proposition~\ref{msr_proof_appendix}}
\label{section_msr_appendix}

\begin{proposition}
	\label{msr_proof_appendix}
For FBDDs and Perceptrons, FN and G-FN can be solved in polynomial time. However, for MLPs, FN can be solved in polynomial time, while solving G-FN is coNP-Complete
\end{proposition}

        As mentioned in Table~\ref{table:results:VerificationResults}, the complexity of \emph{FN} for FBDDs is drawn from the work of~\cite{huang2023feature}. We now move to prove all other complexity classes:

\begin{lemma}
	G-FN can be solved in polynomial time for FBDDs.
\end{lemma}

\emph{Proof.} Let $\langle f, i\rangle$ be an instance. We describe the 
following polynomial algorithm: we enumerate pairs of leaf nodes $(v,v')$ that 
correspond to the paths $(\alpha, \alpha')$. We denote by $\alpha_S$ the subset 
of nodes from $\alpha$ that correspond to the features of $S$. Given the pair 
$(\alpha,\alpha')$ we check if $\alpha$ and $\alpha'$ ``match'' on all features 
from $\{1,\ldots,n\}\setminus \{i\}$ (more formally, there do not exist two 
nodes $v_{\alpha}\in \alpha_{\{1,\ldots,n\}\setminus \{i\}}$ and 
$v_{\alpha'}\in \alpha'_{\{1,\ldots,n\}\setminus \{i\}}$ with the same input 
feature $j$ and with different \emph{path} output edges). If we find two paths $\alpha$ and 
$\alpha'$ that \begin{inparaenum}[(i)]
	\item match on all features in $\{1,\ldots,n\}\setminus \{i\}$, 
	\item \emph{do not} match on feature $i$ (i.e., have different path output 
	edges for input feature $i$), and 
	\item have \emph{the same} label (both classified as True, or both 
	classified as False)
\end{inparaenum} the algorithm returns ``False'' (i.e, $i$ is \emph{not} 
globally necessary with respect to $f$). If we do not encounter any such pair 
$(v,v')$, the algorithm returns ``True''.

Clearly, if the algorithm encounters two paths $(\alpha,\alpha')$ that satisfy 
these three conditions, then it can be concluded that $\{i\}$ is \emph{not} 
contrastive with respect to any assignment $\x$ associated with $\alpha$ and 
$\alpha'$. From Theorem~\ref{necessetiy_contrastive_theorem}, this implies that 
$i$ is not necessary with respect to the corresponding instances of $\langle 
f,\x\rangle$. However, if no such pair was encountered, then there does not 
exist an input $\x\in\mathbb{F}$ for which $\{i\}$ is not contrastive. It hence 
holds that $\{i\}$ is contrastive for any $\langle f,\x\rangle$ and 
Theorem~\ref{necessetiy_contrastive_theorem} thereby implies that $i$ is 
necessary with respect to any $\langle f,\x\rangle$.

\begin{lemma}
	G-FN can be solved in linear time for Perceptrons.
\end{lemma}

\emph{Proof.} Given some $\langle f,i\rangle$ such that $f:=\langle 
\mathbf{w},b\rangle$ is some Perceptron, we can perform a similar process to 
the one described under Lemma~\ref{csr_lemma_perceptrons_appendix} and 
calculate: $max\{\sum_{j\in \{1,\ldots,n\}\setminus \{i\}}y_j\cdot \w_j+b\ | \ 
\y\in \mathbb{F}\}$ as well as: $min\{\sum_{j\in \{1,\ldots,n\}\setminus 
	\{i\}}\y_j\cdot \w_j+b\ | \ \y\in \mathbb{F}\}$ in polynomial time. We now 
	simply 
need to check whether there exists any instance $\x\in\mathbb{F}$ for which:

\begin{equation}
	\begin{aligned}
		\label{some_wierdo_equation}
		\x_i\cdot \w_i+max\{\sum_{j\in \{1,\ldots,n\}\setminus \{i\}}\y_j\cdot 
		\w_j+b\ | \ \y\in \mathbb{F}\}\leq0 \ \ \vee \\
		\x_i\cdot \w_i+min\{\sum_{j\in \{1,\ldots,n\}\setminus \{i\}}\y_j\cdot 
		\w_j+b\ | \ \y\in \mathbb{F}\}>0 
	\end{aligned}
\end{equation}

This condition can obviously be validated in polynomial time since there are 
only two possible relevant scenarios ($\x_i=1$ or $\x_i=0$). If this condition 
holds for one of the two possibilities then there exists an instance 
$\x\in\mathbb{F}$ for which $\{1,\ldots,n\}\setminus \{i\}$ is a local 
sufficient reason of $\langle f,\x\rangle$. It thereby holds that $\{i\}$ is 
\emph{not} a contrastive reason of $\langle f,\x\rangle$. Hence, we can use 
Theorem~\ref{necessetiy_contrastive_theorem}, and conclude that $i$ is not 
necessary with respect to $\langle f,\x\rangle$, thus implying that $i$ is also 
not globally necessary. Should equation~\ref{some_wierdo_equation} not hold, it 
follows that for any $\x\in\mathbb{F}$ the set $\{1,\ldots,n\}\setminus \{i\}$ 
does \emph{not} constitute a local sufficient reason of $\langle f,\x\rangle$. 
This conveys that $\{i\}$ is a local contrastive reason for \emph{any} $\langle 
f,\x\rangle$. Theorem~\ref{necessetiy_contrastive_theorem} further establishes 
that $i$ is necessary for any $\langle f,\x\rangle$, and hence $i$ is 
consequently globally necessary.

	\begin{lemma}
	FN is in PTIME for Perceptrons and MLPs
\end{lemma}

Building upon the correctness of Theorem~\ref{necessetiy_contrastive_theorem}, 
we can deduce that determining the necessity of feature $i$ in relation to 
$\langle f,\x\rangle$ aligns with verifying if $\{i\}$ serves as a contrastive 
reason for $\langle f,\x\rangle$. For both MLPs and Perceptrons, it is possible 
to compute both $f(\x_{\{1,\ldots,n\}\setminus \{i\}};\mathbf{1}_{\{i\}})$ and 
$f(\x_{\{1,\ldots,n\}\setminus \{i\}};\mathbf{0}_{\{i\}})$ and validate whether:

\begin{equation}
	f(\x_{\{1,\ldots,n\}\setminus \{i\}};\mathbf{1}_{\{i\}})\neq 
	f(\x_{\{1,\ldots,n\}\setminus \{i\}};\mathbf{0}_{\{i\}})
\end{equation}

The given condition is satisfied if, and only if, $\{i\}$ is contrastive with 
respect to $\langle f, \mathbf{x} \rangle$, thereby ascertaining whether $i$ is 
necessary in relation to $\langle f, \mathbf{x} \rangle$.

\begin{lemma}
	G-FN is coNP-Complete for MLPs.
\end{lemma}

\emph{Proof.} To obtain membership, given a feature $i$ that we aim to verify 
as globally necessary with respect to $f$, we can guess an instance 
$\x\in\mathbb{F}$ and determine whether:

\begin{equation}
	\label{explanation_definition}
	f(\x_{\{1,\ldots,n\}\setminus \{i\}};\neg\x_{\{i\}})= f(\x)
\end{equation}

In other words, we wish to validate whether fixing all features in 
$\{1,\ldots,n\}\setminus \{i\}$ to their values in $\x$, and negating only the 
value of feature $i$ (to be $\neg \x_i$) changes the prediction of $f(\x)$. 
Clearly, this holds if and only if $\{i\}$ is \emph{not} a contrastive reason 
for $\langle f,\x\rangle$ and from Theorem~\ref{necessetiy_contrastive_theorem} 
this holds if and only if $i$ is \emph{not} necessary with respect to $\langle 
f,\x\rangle$. Put differently, there exists some $\x\in\mathbb{F}$ for which 
$i$ is not necessary with respect to $\langle f,\x\rangle$. This implies that 
$\langle f,i\rangle \not\in$ \emph{G-FN}.

For Hardness, we will make use of the following Lemma whose proof appears in 
the work of~\cite{BaMoPeSu20}.

\begin{lemma}
	\label{lemma_boolean_circuit}
	If we have a Boolean circuit $B$, we can create an MLP $f_{B}$ in 
	polynomial time that represents an equivalent Boolean function with respect 
	to $B$.
\end{lemma}

We now prove hardness by reducing from \emph{TAUT}, a well-known coNP-Complete 
problem which is defined as follows:

\noindent\fbox{%
	\parbox{\textwidth}{%
		\mysubsection{\textbf{TAUT} (Tautology)}:
		
		\textbf{Input}: A boolean formula $\psi$ 
		
		\textbf{Output}: \yes{}, if $\psi$ is a tautology and \no{} otherwise.
	}%
}

Given some $\langle \psi\rangle$ with variables: $x_1,\ldots x_n$ we introduce a new variable $x_{n+1}$ and 
construct a new boolean formula:

\begin{equation}
	\psi':=\psi \wedge x_{n+1}
\end{equation}

We then can use Lemma~\ref{lemma_boolean_circuit} to transform it to an MLP $f$ 
and construct $\langle f,i:=n+1\rangle$. 

If $\langle \psi \rangle\in$ \emph{TAUT} then it holds that $\psi$ is always True, and hence:

\begin{equation}
	f(\x_{\{1,\ldots,n\}};\mathbf{1}_{\{n+1\}})= 1 \ \ \wedge \ \ 
	f(\x_{\{1,\ldots,n\}};\mathbf{0}_{\{n+1\}})= 0
\end{equation}

where $\mathbf{1}_{\{n+1\}}$ and $\mathbf{0}_{\{n+1\}}$ denote that feature $n+1$ is 
set to either $1$ or $0$.

Hence, for any value $\x\in\mathbb{F}$ we can find a corresponding instance 
$\z\in\mathbb{F}$ such that:
\begin{equation}
	f(\x_{\{1,\ldots,n\}};\z_{\{n+1\}})\neq f(\x)
\end{equation}

This implies that the subset $\{n+1\}$ is contrastive with respect to any 
$\langle f,\x\rangle$ and from theorem~\ref{necessetiy_contrastive_theorem}, feature 
$n+1$ is necessary with respect to any $\langle f,\x\rangle$. Thus, it 
satisfies that feature $n+1$ is globally necessary with respect to $f$ and 
consequently, $\langle f,i\rangle \in$ \emph{G-FN}.

Let us now consider the scenario where $\langle \psi \rangle\not\in$ 
\emph{TAUT}. Under this assumption, it follows that there exists a False 
assignment for $\langle x_1,\ldots,x_n\rangle$, rendering $\psi'$ false 
irrespective of the assignment to $\x_{n+1}$. To put it differently, this 
scenario satisfies the following condition:

\begin{equation}
	f(\x_{\{1,\ldots,n\}};\mathbf{1}_{\{n+1\}})= 0 \ \ \wedge \ \ 
	f(\x_{\{1,\ldots,n\}};\mathbf{0}_{\{n+1\}})= 0
\end{equation}

Thus, we can take an arbitrary vector $\x$ and set some other arbitrary vector 
$\z$ to be equal to $\x$ on the first $n$ features and negated on feature 
$n+1$. Both of these vectors will be labeled to class $0$, hence:

\begin{equation}
	\exists \z, \x\in \mathbb{F} \quad f(\x_{\{1,\ldots,n\}\setminus 
	\{i\}};\z_{\{n+1\}})=f(\x_{\{1,\ldots,n\}\setminus 
	\{i\}};\neg\z_{\{n+1\}})=0= f(\x)
\end{equation}

We can thus conclude that $\{n+1\}$ is \emph{not} a contrastive reason of 
$\langle f,\x\rangle$ and from theorem~\ref{necessetiy_contrastive_theorem}, 
this implies that $n+1$ is not necessary with respect to $\langle f,\x\rangle$. 
Particularly, $n+1$ is not globally necessary, consequently implying that 
$\langle f,i\rangle \not\in$ \emph{G-FN}.

\section{Proof of Proposition~\ref{fn_complexity_appendix}}
\label{section_fn_appendix}

\begin{proposition}
	\label{fn_complexity_appendix}
    For FBDDs, G-CSR and G-MSR can be solved in polynomial time, while MSR is NP-Complete. Moreover, in MLPs, solving G-CSR and G-MSR is coNP-Complete, while solving MSR is $\Sigma^P_2$-Complete.
\end{proposition}

        As mentioned in Table~\ref{table:results:VerificationResults}, the complexity of \emph{MSR} for FBDDs and \emph{MSR} for MLPs are drawn from the work of~\cite{BaMoPeSu20}. We now move to prove all other complexity classes:

\begin{lemma}
	G-CSR can be solved in polynomial time for FBDDs.
\end{lemma}
\emph{Proof.} Let $\langle f, S\rangle$ be an instance. We describe the 
following polynomial algorithm: We enumerate pairs of leaf nodes $(v,v')$ that 
correspond to the paths $(\alpha, \alpha')$. Let us denote by $\alpha_S$ the 
subset of nodes from $\alpha$ that correspond to the features of $S$. Given the 
pair $(\alpha,\alpha')$, the algorithm checks if $\alpha$ and $\alpha'$ 
``match'' on all features from $S$ (more formally, there do not exist two nodes 
$v_{\alpha}\in \alpha_S$ and $v_{\alpha'}\in \alpha'_{S}$ with the same input 
feature $i$ and with different path output edges). If we find two paths $\alpha$ and 
$\alpha'$ that match all features in $S$, and that have different labels 
(one classified as True and the other: False) the algorithm returns ``False'' 
(i.e., $S$ is not a global sufficient reason of $f$). If we do not encounter 
any such pair $(v,v')$, the algorithm returns True.

\begin{lemma}
	G-MSR is in PTIME for FBDDs.
\end{lemma}

\emph{Proof.} Since \emph{G-CSR} is in PTIME for FBDDs, we can use 
Proposition~\ref{any_ordering_converges_appendix} which states that 
algorithm~\ref{alg:subset-minimal-global} always converges to the unique global 
cardinally minimal sufficient reason after a linear number of calls checking 
whether $\mySuff(f,S\setminus\{i\})=1$. Each one of these calls can be 
performed in polynomial time (since \emph{G-CSR} is polynomial for FBDDs), so 
hence using algorithm~\ref{alg:subset-minimal-global}, we can obtain the unique 
global cardinally minimal sufficient reason of $f$ in polynomial time, and 
return True if its size is smaller or equal to $k$, and False otherwise.

\begin{lemma}
	G-CSR is coNP-Complete for MLPs.
\end{lemma}

\emph{Proof.} Membership is straightforward and is obtained since we can guess 
some $\x,\z\in\mathbb{F}$ and validate whether it satisfies that 
$f(\x_{S};\z_{\Bar{S}}) \neq f(\x)$. If so, $\langle f, S\rangle\not\in$ 
\emph{G-CSR}.

Given our forthcoming proof that the \emph{G-CSR} query for Perceptrons is 
coNP-Hard, it follows straightforwardly that the same is true for MLPs.  
Nevertheless, we show how hardness can also be proved particularly for MLPs via 
a reduction from the (local) \emph{CSR} explainability query for MLPs.

Given the tuple $\langle f,\x,S\rangle$ we construct an MLP $f'$ which 
satisfies the following conditions:

\begin{equation}
	f'(\y)=\begin{cases}
		f(\y) \quad if \ \ (\x_S= \y_S)\\
		1 \quad if \ \ (\x_S\neq \y_S)
	\end{cases}
\end{equation}

An MLP corresponding to this specification can be built by applying Lemma~\ref{lemma_boolean_circuit}, which asserts that any boolean circuit can be represented as an equivalent MLP. Specifically, we can encode the condition $(\x_i\oplus \y_i)$ for each $i\in S$ and define:

\begin{equation}
    \psi:= \bigwedge_{i\in S} (\x_i\oplus y_i)
\end{equation}

Using Lemma~\ref{lemma_boolean_circuit}, we can convert $\psi$ into an MLP $g$ and subsequently integrate $g$ with the input layer of the original MLP, $f$. This configuration results in the input layer of $f$ receiving connections from two distinct MLPs, each generating different outputs. Further applying Lemma~\ref{lemma_boolean_circuit}, we introduce an additional hidden layer that produces a single output representing the disjunction of the two preceding outputs. This newly constructed MLP corresponds to the structure of $f'$. To preserve the integrity of the MLP's architecture, we can implement zero-weights and zero-biases for any unconnected neuron connections.

If $\langle f,\x,S\rangle \in$ \emph{CSR}, then it satisfies that:

\begin{equation}
	\label{explanation_definition}
	\forall(\z\in \mathbb{F}).\quad [f(\x_{S};\z_{\Bar{S}})=f(\x)]
\end{equation}

Given that $f'(\y)=f(\y)$ holds for any input for which $\x_S=\y_S$, then it also 
satisfies that: 

\begin{equation}
	\begin{aligned}
		\label{explanation_definition}
		\forall(\z\in \mathbb{F}).\quad [f'(\x_{S};\z_{\Bar{S}})=f'(\x)] \iff \\
		\forall(\x, \z\in \mathbb{F}).\quad (\x_S= \z_S)\to 
		[f'(\x_{S};\z_{\Bar{S}})=f'(\x)]
	\end{aligned}
\end{equation}

If $\x_S\neq \y_S$ then it consequently holds that $f'(\y)=1$. This implies that:

\begin{equation}
	\label{explanation_definition}
	\forall(\x, \z\in \mathbb{F}).\quad (\x_S\neq \z_S)\to 
	[f'(\x_{S};\z_{\Bar{S}})=f'(\x)=1]
\end{equation}

Overall, we arrive at that:

\begin{equation}
	\label{explanation_definition}
	\forall(\x, \z\in \mathbb{F}). \quad [f'(\x_{S};\z_{\Bar{S}})=f'(\x)]
\end{equation}

implying that $\langle f',S\rangle\in$ \emph{G-CSR}.

If $\langle f,\x,S\rangle \not\in$ \emph{CSR}, then it satisfies that:
\begin{equation}
	\label{explanation_definition}
	\exists(\z\in \mathbb{F}).\quad [f(\x_{S};\z_{\Bar{S}})\neq f(\x)]
\end{equation}

Given that $f'(\y)=f(\y)$, it follows that for any input satisfying $\x_S=\y_S$ 
the following condition is also met: 

\begin{equation}
	\begin{aligned}
		\label{explanation_definition}
		\exists(\z\in \mathbb{F}).\quad [f'(\x_{S};\z_{\Bar{S}})\neq f'(\x)]
	\end{aligned}
\end{equation}

implying that:

\begin{equation}
	\begin{aligned}
		\exists(\x, \z\in \mathbb{F}). \quad [f'(\x_{S};\z_{\Bar{S}})\neq 
		f'(\x)]
	\end{aligned}
\end{equation}

Thus, it holds that $\langle f',S\rangle\not\in$ \emph{G-CSR}.

\begin{lemma}
	G-MSR is coNP complete for MLPs.
\end{lemma}
Both hardness and membership results trivially derive from those described for 
Perceptrons.
	
	\section{Proof of Proposition~\ref{cc_complexity_appendix}}
	\label{section_cc_appendix}
	
	\begin{proposition}
		\label{cc_complexity_appendix}
	 For FBDDs, G-FR can be solved in polynomial time, while solving FR is 
	 coNP-Complete. Moreover, in MLPs, solving G-FR is coNP-Complete, while 
	 solving FR is $\Pi^P_2$-Complete.
	\end{proposition}

         As mentioned in Table~\ref{table:results:VerificationResults}, the complexity of \emph{MSR} for MLPs is drawn from the work of~\cite{huang2021efficiently} which proved hardness for DNF classifiers (and this also holds for MLPs, from Lemma~\ref{lemma_boolean_circuit}). Moreover, the complexity of \emph{FR} for FBDDs is drawn from the work of~\cite{huang2023feature}. We now move to prove all other complexity classes:

	\begin{lemma}
		G-FR can be solved in polynomial time for FBDDs.
	\end{lemma}
	
	\emph{Proof.} Let $\langle f, i\rangle$ be an instance. We describe the 
	following polynomial algorithm: We enumerate pairs of leaf nodes $(v,v')$ 
	that 
	correspond to the paths $(\alpha, \alpha')$. We denote by $\alpha_S$ the 
	subset 
	of nodes from $\alpha$ that correspond to the features of $S$. Given the 
	pair 
	$(\alpha,\alpha')$ we check if $\alpha$ and $\alpha'$ ``match'' on all 
	features 
	from $\{1,\ldots,n\}\setminus \{i\}$ (more formally, there do not exist two 
	nodes $v_{\alpha}\in \alpha_{\{1,\ldots,n\}\setminus \{i\}}$ and 
	$v_{\alpha'}\in \alpha'_{\{1,\ldots,n\}\setminus \{i\}}$ with the same 
	input 
	feature $j$ and with different output edges). If we find two paths $\alpha$ 
	and 
	$\alpha'$ that \begin{inparaenum}[(i)]
		\item match on all features in $\{1,\ldots,n\}\setminus \{i\}$, 
		\item \emph{do not} match on feature $i$ (i.e., have different output 
		edges), and 
		\item have different labels (one is classified as True and the other: 
		False)
	\end{inparaenum} the algorithm returns ``False'' (i.e, $i$ is \emph{not} 
	redundant with respect to $f$). If we do not encounter any such pair 
	$(v,v')$, 
	the algorithm returns ``True''.

	\begin{lemma}
		G-FR is coNP-Complete for MLPs.
	\end{lemma}
	
	Both hardness and membership proofs for Perceptrons (proved in the 
	following section) also trivially hold for 
	MLPs.

	\section{Proof of Proposition~\ref{local_complexity_appendix}}
	\label{section_local_proofs_appendix}
				
				\begin{proposition}
					\label{local_complexity_appendix}
				    For Perceptrons, solving FR and G-FR are both coNP-Complete.
				\end{proposition}

				\begin{lemma}
					FR is coNP-Complete for Perceptrons
				\end{lemma}
				
				\emph{Proof.} \textbf{Membership.} We recall that validating whether a subset $S$ is a local sufficient reason with respect to some $\langle f,\x\rangle$ can be done in polynomial time for Perceptrons, as was elaborated on in Lemma~\ref{csr_lemma_perceptrons_appendix}. This can be done by polynomially calculating both: $max\{\sum_{j\in \Bar{S}}\y_j\cdot \w_j+b\ | \ \y\in \mathbb{F}\}$ and $min\{\sum_{j\in \Bar{S}}\y_j\cdot \w_j+b\ | \ \y\in \mathbb{F}\}$ and then validating whether it holds that:

				\begin{equation}
					\begin{aligned}
						\x_i\cdot \w_i+max\{\sum_{j\in \Bar{S}}\y_j\cdot \w_j+b\ | \ \y\in \mathbb{F}\}\leq0 \ \ \vee \\
						\x_i\cdot \w_i+max\{\sum_{j\in \Bar{S}}\y_j\cdot w_j+b\ | \ \y\in \mathbb{F}\}>0 
					\end{aligned}
				\end{equation}
				Hence, membership in coNP holds since we can guess some subset $S\subseteq \{1,\ldots,n\}$ and polynomially validate whether it holds that:
				
				\begin{equation}
					\begin{aligned}
						\mySuff(f,\x,S)=1 \ \ \wedge \ \ \mySuff(f,\x,S\setminus \{i\})=0
					\end{aligned}
				\end{equation}

				If the following condition holds, then it satisfies that $i$ is not redundant with respect to $\langle f,\x\rangle $ and hence $\langle f,i\rangle \not\in$ \emph{FR}.
				
				\textbf{Hardness.} We reduce \emph{FR} for Perceptrons from the subset sum problem (SSP), specifically from $\overline{SSP}$ which is coNP-Complete. Given some $\langle (\z_1,\z_2,\ldots,\z_n),T\rangle$ construct a Perceptron $f:=\langle \w,b\rangle$ where we set $\w:=(\z_1,\z_2,\ldots,\z_n)\cdot (\w_{n+1})$ ($\w$ is of size $n+1$), where $\w_{n+1}:=\frac{1}{2}$, and $b:=-(T+\frac{1}{4})$. The reduction computes $\langle f,\x:=\mathbf{1}, i:=n+1\rangle$.
				
				Assume that $\langle (\z_1,\z_2,\ldots,\z_n),T\rangle \in \overline{SSP}$. This implies that there does not exist any subset $S\subseteq\{1,\ldots, n\}$ for which $\sum_{j\in S}\z_j = T$. Given that the values in $(\z_1,\ldots, \z_n)$ are \emph{integers}, it consequently follows that there does not exist a subset $S$ satisfying that:
				
				\begin{equation}
					\begin{aligned}
						T-\frac{1}{2}< \sum_{j\in S}\z_j < T+\frac{1}{2}
					\end{aligned}
				\end{equation}
				
				Consequently, it holds that there is no subset $S\subseteq \{1,\ldots,n\}$ for which:
				
				\begin{equation}
					\begin{aligned}
						T-\frac{1}{2}<\sum_{j\in S}\mathbf{w}_j \cdot 1  < T+\frac{1}{2} \iff \\
						-\frac{3} {4} <\sum_{j\in S}\mathbf{w}_j \cdot 1 + b  < \frac{1}{4}				
					\end{aligned}
				\end{equation}
				
				which is equivalent to saying that for any subset $S\subseteq \{1,\ldots,n\}$ it holds that:
    		\begin{equation}
					\begin{aligned}
						[ \ \sum_{j\in S}\mathbf{w}_j \cdot 1 + b  < -\frac{3} {4}  \ ] \ \ \vee \ \ 				[ \ \sum_{j\in S}\mathbf{w}_j \cdot 1 + b  > \frac{1} {4}  \ ] 
					\end{aligned}
				\end{equation}

This implies that:

            		\begin{equation}
					\begin{aligned}
					\forall S'\subseteq \{1,\ldots,n+1\} \ \ [ \	[ \ \sum_{j\in S}\mathbf{w}_j \cdot 1 +\w_{n+1}\cdot 0+ b  < 0  \ ] \ \ \wedge \ \
     [ \ \sum_{j\in S}\mathbf{w}_j \cdot 1 +\w_{n+1}\cdot 1+ b  < -\frac{1} {4}  \ ] \ ] \ \ \vee \\
[ \ [ \ \sum_{j\in S}\mathbf{w}_j \cdot 1 +\w_{n+1}\cdot 0+ b  > \frac{1} {4}  \ ] \ ] \ \ \wedge  \ \ 				[ \ \sum_{j\in S}\mathbf{w}_j \cdot 1 +\w_{n+1}\cdot 1+ b  > \frac{3} {4}  \ ] \ ]
					\end{aligned}
				\end{equation}

Thus, if we set the values of all features $\{1,\ldots,n+1\}$ to $1$, modifying the value of feature $n+1$ from $1$ to $0$ will not alter the classification (as $f$ will continue to be either positive or negative). Equivalently:

        		\begin{equation}
					\begin{aligned}
\forall S'\subseteq \{1,\ldots,n\} \ \ \ [ \ [ \ f(\mathbf{1}_{S'};\mathbf{0}_{\{n+1\}})=0 \ \ \wedge \ \  [ \  f(\mathbf{1}_{S'};\mathbf{1}_{\{n+1\}})=0 \ ] \ ] \ \vee \\ [ \ [ \ f(\mathbf{1}_{S'};\mathbf{0}_{\{n+1\}})=1 \ \ \wedge \ \   
[ \ f(\mathbf{1}_{S'};\mathbf{1}_{\{n+1\}})=1 \ ] \ ]
					\end{aligned}
				\end{equation}
    
 In other words, it can be stated that:

				\begin{equation}
					\begin{aligned}
						\forall S'\subseteq \{1,\ldots, n, n+1\}\quad  \mySuff(f,\mathbf{1},S')=1\to  \mySuff(f,\mathbf{1},S'\setminus \{n+1\})=1
					\end{aligned}
				\end{equation}
				
				Therefore, feature $n+1$ is redundant with respect to $\langle f,\mathbf{1}\rangle$, implying that $\langle f,\mathbf{1},i\rangle\in$ \emph{FR}.
				
				Let us assume that $\langle (\z_1,\z_2,\ldots,\z_n),T\rangle \not\in \overline{SSP}$. From this assumption, it follows that there exists a subset of features, $S\subseteq \{z_1,\ldots \z_n\}$ for which: $\sum_{j\in S}\z_j = T$. We can express this equivalently as:

				\begin{equation}
					\begin{aligned}
						\label{last_relevant_equation}
						T = \sum_{j\in S}\z_j \iff
						-\frac{1}{4} = \sum_{j\in S}\mathbf{w}_j + b \iff \\
						[-\frac{1}{4} = \sum_{j\in S}\mathbf{w}_j+ \mathbf{w}_{n+1}\cdot 0+ b] \ \ \wedge \ \ 		[\frac{1}{4} = \sum_{j\in S}\mathbf{w}_j+ \mathbf{w}_{n+1}\cdot 1+ b]
					\end{aligned}
				\end{equation}
				
				We denote $S':=S\cup \{n+1\}$. Based on equation~\ref{last_relevant_equation}, we have that $f(\mathbf{1}_{S'};\mathbf{0}_{\Bar{S'}})>0$. Moreover, given that all features in $\Bar{S'}$ are positive integers, then if we add additional features to $S'$ (making it larger) the classification will necessarily remain positive. More formally, it is also established that for any $S''\subseteq \{1,\ldots, n+1\}$ for which $S'\subseteq S''$ the following holds: $f(\mathbf{1}_{S''};\mathbf{0}_{\Bar{S''}})>0$. Hence, fixing the features of $S'$ to value $1$ maintains a positive value for $f$ (and hence a $1$ classification), and hence $S'$ is sufficient with respect to $\langle f,\mathbf{1}\rangle$. Referring to equation~\ref{last_relevant_equation}, we observe that: $f(\mathbf{1}_{S'\setminus \{n+1\}};\mathbf{0}_{\Bar{S'}\cup \{n+1\}})=0$. This implies that $S'\setminus \{n+1\}$ is \emph{not} sufficient with respect to $\langle f,\mathbf{1}\rangle$. In other words, we can conclude that:
				
				\begin{equation}
					\begin{aligned}
						\exists S'\subseteq \{1,\ldots, n, n+1\}\quad  \mySuff(f,\mathbf{1},S')=1\wedge  \mySuff(f,\mathbf{1},S'\setminus \{n+1\})=0
					\end{aligned}
				\end{equation}
				
				Consequently, feature $n+1$ is \emph{not} redundant with respect to $\langle f,\mathbf{1}\rangle$, thus implying that $\langle f,\x,i\rangle\not\in$ \emph{FR}.

\begin{lemma}
	G-FR is coNP-Complete for Perceptrons.
\end{lemma}

\emph{Proof.} Membership is established from the fact that one can guess 
some 
$\x,\z\in\mathbb{F}$ and validate whether: $f(\x_{\{1,\ldots,n\}\setminus 
	\{i\}};\z_{\{i\}})\neq f(\x)$. From 
Theorem~\ref{necessitiy_contrastive_iff}, 
this condition holds if and only if $i$ is necessary with respect to 
$\langle 
f,\x\rangle$. Furthermore, 
Proposition~\ref{split_necessary_redundant_appendix} 
establishes that this situation is equivalent to $i$ being \emph{not} 
globally 
redundant with respect to $f$, thereby implying $\langle f,i\rangle\not\in$ 
\emph{G-FR}.

Before proving hardness, we will make use of the following Lemma which is 
simply a refined version of 
Proposition~\ref{last_proposition_uniquenss_appendix}:

\begin{lemma}
	\label{fr_helper_appendix}
	$S$ is a global sufficient reason of $f$ iff for any $i\in 
	\overline{S}$, 
	$i$ is globally redundant.
\end{lemma}

\emph{Proof.} $S$ is a global sufficient reason of $f$ if and only if there 
exists some $S'\subseteq S$ which is a subset minimal global sufficient 
reason 
of $f$. From Proposition~\ref{last_proposition_uniquenss_appendix}, it 
holds 
that any feature $i\in \Bar{S'}$ is globally redundant, and since 
$\Bar{S}\subseteq \Bar{S'}$, it satisfies that any feature $i\in \bar{S}$ 
is 
globally redundant.

We are now in a position to employ Lemma~\ref{fr_helper_appendix}, from 
which 
we can discern that $S$ qualifies as a global sufficient reason of $f$ if 
and 
only if every $i\in \Bar{S}$ is globally redundant. Consequently, we can 
leverage the reduction that was utilized for establishing the coNP-Hardness 
of 
\emph{G-CSR} for Perceptrons, as detailed in 
Lemma~\ref{csr_lemma_perceptrons_appendix}.

In other words, we can reduce \emph{G-FR} for Perceptrons from 
$\overline{SSP}$. Given some $\langle (\z_1,\z_2,\ldots,\z_n),T\rangle$ we can 
again construct a Perceptron $f:=\langle \mathbf{w},b\rangle$ where 
$\mathbf{w}:=(\z_1,\z_2,\ldots,\z_n)\cdot (\mathbf{w}_{n+1})$ ($\mathbf{w}$ is 
of 
size $n+1$), $\mathbf{w}_{n+1}:=\frac{1}{2}$, and $b:=-(T+\frac{1}{4})$. 
The 
reduction computes $\langle f,i:=n+1\rangle$. 

It has been established in Lemma~\ref{csr_lemma_perceptrons_appendix} that 
$S=\{1,2,\ldots,n\}$ serves as a  global sufficient reason of $f$ if and 
only 
if no subset $S'\subseteq S=\{1,2,\ldots,n\}$ exists for which $\sum_{i\in 
	S'}\z_i=T$. Moreover, due to Lemma~\ref{fr_helper_appendix}, the set 
$S=\{1,2,\ldots,n\}$ is a global sufficient reason of $f$ if and only if 
any 
feature in $\Bar{S}$ is globally redundant. However, $\Bar{S}$  is in our 
case 
simply $\{n+1\}$. This leads to the conclusion that feature $n+1$ is 
globally 
redundant, thereby concluding the reduction.

\section{Proof of Proposition~\ref{fr_proof_appendix}}
\label{section_fr_appendix}

\begin{proposition}
	\label{fr_proof_appendix}
	For FBDDs, both CC and G-CC can be solved in polynomial time. Moreover, for 
	Perceptrons and MLPs, solving both CC or G-CC is $\#P$-Complete.
\end{proposition}

\begin{lemma}
	\label{cc_complexity_appendix_perceptrons}
	G-CC is $\#P$-Complete for Perceptrons.
\end{lemma}

For simplification, we follow common conventions~\cite{BaMoPeSu20} and 
prove that the global counting procedure for: $C(S,f)=|\{\x\in 
\mathbb{F},\z\in \{0,1\}^{\vert\overline{S}\vert}, 
f(\x_{S};\z_{\Bar{S}}) = f(\x)\}|$ is $\#P$-Complete, rather than $c(S,f)$. 
Clearly, it holds that: $C(S,f)=c(S,f)\cdot 2^{|\Bar{S}|+n}$ and hence $c(S,f)$ 
and $C(S,f)$ are interchangeable. 
We observe that computing $2^{|\Bar{S}|+n}$ can be executed in polynomial time because $n$, which denotes the number of input features, is given in \emph{unary} (the full size of the input encoding is at least $n$).

\textbf{Membership.} Membership is straightforward since the sum: $|\{\x\in 
\mathbb{F},\z\in \{0,1\}^{\vert\overline{S}\vert}, 
f(\x_{S};\z_{\Bar{S}}) = f(\x)\}|$ is equivalent to the sum of certificates 
$(\x,\z)$ satisfying:

\begin{equation}
	\exists\x\in \mathbb{F},\exists\z\in \{0,1\}^{\vert\overline{S}\vert}, 
	f(\x_{S};\z_{\Bar{S}}) = f(\x)
\end{equation}

which is of course polynomially verifiable.

%Membership in $\#P$ is trivial, as one can count in $\#P$ the all assignments 
%of the type $(\x,\z,S)$, such that they saisfy: 
%$\{\x\in \mathbb{F},\z\in \{0,1\}^{\vert\overline{S}\vert}, 
%f(\x_{S};\z_{\Bar{S}}) = f(\x)\}$. It is straightforward to show that 
%chrecking 
%the 
%validity of each satisfying assignemnt is in NP, as a nondetermenistic 
%Turing Machine can check the validity of each assignment in polynomial time. 
%The sum of all such solutions allows the computation (in polynomial time) of 
%$c(S,f)$, by dividing the accumulated by $\vert \mathbb{F} \vert \cdot \vert 
%\{0,1\}^{\vert 
%	\overline{S} \vert}\vert=2^{n}\cdot 2^{n-\vert S \vert } = 2^{2n-\vert S 
%	\vert}$.
%Hence, it follows directly that this counting task is contained in $\#P$.

\textbf{Hardness.} We reduce from (local) \emph{CC} of Perceptrons, which is 
$\# P$-Complete~\cite{BaMoPeSu20}. Given some $\langle f,S,\x\rangle$, where $f:=\langle 
\mathbf{w},b\rangle$ is a Perceptron, the reduction computes $f(\x)$ and if 
$f(\x)=1$ constructs $f':=\langle \mathbf{w}',b'\rangle$ such that $b':= 
b+\sum_{i\in 
	S}(\x_{i}\cdot \w_{i})$, and $\mathbf{w'}:= 
(\mathbf{w}_{
	\Bar{S}},\delta)$, with 
$\delta:=(\sum_{i\in \overline{S}}\vert w_{i}\vert) - b' $. $\mathbf{w}_{
	\Bar{S}}$ denotes a partial assignment where all features of the subset 
$\Bar{S}$ are drawn from the vector $\mathbf{w}$ (the vector $\mathbf{w'}$ 
is of size $|\Bar{S}|+1$). If $f(\x)=0$, the reduction constructs 
$f':=\langle \mathbf{w}',b'\rangle$ with the same $b'$ but with 
$\mathbf{w'}:= 
(\mathbf{w}_{
	\Bar{S}},\delta')$, such that 
$\delta':=-(\sum_{i\in \overline{S}}\vert \w_{i}\vert) - b'-1$.

For both reduction scenarios ($f(\x)=1$ or $f(\x)=0$) we will demonstrate that 
given the \emph{global} completion count $C(\emptyset, f')$ we can determine 
the \emph{local} completion count of $c(S,f,\x)$ in polynomial time. We do this 
by proving the following Lemma:

\begin{lemma}
	Given the polynomial construction of $f'$, it satisfies that:
	\begin{equation} 
		C(S,f,\x) = \sqrt{\frac{1}{2}\cdot 
			C(\emptyset,f')-2^{2\vert \overline{S}\vert}}
	\end{equation}
\end{lemma}

If this lemma is proven, then demonstrating the hardness of the global completion count for perceptrons becomes straightforward. This is because $n$, representing the number of input features, is given in unary. Consequently, computations such as $2^{O(n)}$ (and therefore $2^{2\vert \overline{S}\vert}$) can be performed in polynomial time. Additionally, since the \emph{binary} representation of $[\frac{1}{2}\cdot 
			C(\emptyset,f')-2^{2\vert \overline{S}\vert}]$ is of size $O(n)$ ($C(\emptyset,f')$ is bounded by $2^{2n}$), computing the square root of this expression can also be achieved in polynomial time.

\emph{Proof of Lemma.} We denote $m$ and $t$ as the \emph{number} of assignments  $\x' \in 
\{0,1\}^{\vert \overline{S}\vert + 1}$, for which $f'$ predicts 0 or 1. 
Formally:

\begin{align}
	m & \coloneqq \left \vert \left\{ \x'\in \{0,1\}^{\vert\overline{S}\vert 
		+1} \, \middle\vert \, f'(\x')=1 \right\} \right \vert \ \ , \ \
	t  \coloneqq \left \vert \left\{ \x'\in \{0,1\}^{\vert\overline{S}\vert 
		+1} \, \middle\vert \, f'(\x')=0 \right\} \right \vert
\end{align}

Clearly, it holds that:

\begin{equation}
	\label{part1_proof_addition}
	m + t = 2^{|\overline{S}|+1}
\end{equation}

It also satisfies that:

\begin{equation}
    	\label{part2_proof_addition}
					\begin{aligned}
	C(S:=\emptyset,f')= |\{\x'\in \{0,1\}^{|\overline{S}|+1},\z\in 
	\{0,1\}^{\vert\overline{S}\vert+1}, 
	f'(\x'_{S};\z_{\Bar{S}}) = f'(\x')\}|= \\
 |\{\x'\in \{0,1\}^{|\overline{S}|+1},\z\in 
	\{0,1\}^{\vert\overline{S}\vert+1}, 
	f'(\z) = f'(\x')\}|
 =m^2+t^2
 \end{aligned}
\end{equation}

This is true because there are precisely $m^2$ pairs $(\x',\z)$ where both $f'(\z)=1$ and $f'(\x')=1$, and exactly $t^2$ pairs where $f'(\z)=0$ and $f'(\x')=0$. As a result of equations~\ref{part1_proof_addition} 
and~\ref{part2_proof_addition}, it satisfies that:

\begin{equation}
	C(\emptyset,f')= m^2+(2^{|\Bar{S}|+1}-m)^2
\end{equation}

This implies that the aforementioned values of $m/t$ obey the following 
quadratic relation:

\begin{align}
	m/t &= \frac{-(-2^{\vert \overline{S}\vert + 2})\pm\sqrt{(-2^{\vert 
				\overline{S}\vert + 2})^2 - 4\cdot 2 \cdot (2^{2 \vert 
				\overline{S}\vert+2} 
			-C(\emptyset,f'))}}{2\cdot 2} \nonumber \\
	&= \frac{2^{\vert \overline{S}\vert + 2}\pm\sqrt{2^{2\vert 
				\overline{S}\vert + 4} - 8\cdot (2^{2 \vert 
				\overline{S}\vert+2} 
			-C(\emptyset,f'))}}{4} \nonumber \\
	&= 2^{\vert \overline{S} \vert} \pm \sqrt{2^{2\vert\overline{S}\vert}-(2^{2 
				\vert \overline{S} \vert+1} -\frac{1}{2}\cdot 
			C(\emptyset,f'))} \nonumber
\end{align}

Accordingly, $m/t$ must obey the following condition:

\begin{equation} 
	m/t = 2^{\vert \overline{S} \vert} \pm \sqrt{\frac{1}{2}\cdot 
		C(\emptyset,f')-2^{2\vert \overline{S}\vert}}
\end{equation}

We observe that one of the terms exceeds $2^{|\overline{S}|}$, while the other is less than $2^{|\overline{S}|}$. Therefore, to ascertain whether $m$ or $t$ corresponds to the first or second term, it suffices to compare their counts to $2^{|\overline{S}|}$. We will start by proving the first part of the Lemma. Specifically, to 
establish that when $f(\x)=1$ the following condition is satisfied: 
\begin{equation}
	C(S,f,\x) = \sqrt{\frac{1}{2}\cdot 
		C(\emptyset,f')-2^{2\vert \overline{S}\vert}}
\end{equation}

We will first prove that when $f(\x)=1$, then there are at least 
$2^{|\Bar{S}|}$ vectors $\x'\in\{0,1\}^{|\Bar{S}|+1}$ for which $f'(\x')=1$. 

First, we assume that $\x'_{|\Bar{S}|+1}=1$ (the assignment for feature $|\Bar{S}|+1$ is $1$). For all $\x'\in\{0,1\}^{|S|+1}$ 
such that $\x'_{|\Bar{S}|+1}=1$ it holds that:

\begin{equation}
	\begin{aligned}
		\w'\cdot \x' + b' 
		=\\
		\w'_{\Bar{S}}\cdot \x'_{\Bar{S}}
		+ \w'_{\vert\overline{S}\vert+1} \cdot 
		\x'_{\vert\overline{S}\vert+1}
		+ b'
		=\\
		\w'_{\Bar{S}}\cdot \x'_{\Bar{S}}
		+  \delta
		+ b'		
		=\\
		\w'_{\Bar{S}}\cdot \x'_{\Bar{S}} 
		+ \left ((\sum_{i\in \overline{S}}\vert \w_{i}\vert) - b'\right ) 
		+ b'		
		=\\
		\w'_{\Bar{S}}\cdot \x'_{\Bar{S}} 
		+ \sum_{i\in \overline{S}}\vert \w_{i}\vert
		\geq 0
	\end{aligned}
\end{equation}

Given that there are precisely $2^{|\overline{S}|}$ assignments where 
$\x'_{|\overline{S}|+1}=1$, it can be inferred that there are at least 
$2^{|\overline{S}|}$ assignments for which $f'(\x')=1$. Hence, the following 
condition holds:

\begin{equation}
\label{equation_to_cite_now_what}
	m =
	2^{\vert \overline{S} \vert} + \sqrt{\frac{1}{2}\cdot 
		C(\emptyset,f')-2^{2\vert \overline{S}\vert}}
\end{equation}

The count in equation~\ref{equation_to_cite_now_what} corresponds to the number of positive assignments for which $f(\x')=1$. As mentioned, out of these, there are $2^{|\overline{S}|}$ assignments where $\x'_{|S|+1}=1$. Consequently, the exact number of assignments with $\x'_{|S|+1}=0$ that satisfy 
that $f'(\x)=1$ is exactly:

\begin{equation} 
	( \ 2^{\vert \overline{S} \vert}+\sqrt{\frac{1}{2}\cdot 
		C(\emptyset,f')-2^{2\vert \overline{S}\vert}} \ )-2^{\vert \overline{S} \vert}=\sqrt{\frac{1}{2}\cdot 
		C(\emptyset,f')-2^{2\vert \overline{S}\vert}}
\end{equation}

Furthermore, it holds that:

\begin{equation} 
	\begin{aligned}
		f'(\x'_{\Bar{S}};\mathbf{0}_{|\Bar{S}|+1})= 
		step( \ \w'_{\Bar{S}}\cdot \x'_{\Bar{S}}+ 0 + b' \ ) 
		=\\
		step( \ \w_{\Bar{S}}\cdot \x'_{\Bar{S}} +
		b+\sum_{i\in 
			S}(\x_{i}\cdot \w_{i}) \ )
		=\\
		step( \ \w_{\Bar{S}}\cdot \x'_{\Bar{S}} +
		w_{S}\cdot \x_{S} + b \ )
		= f(\x_S;\x'_{\Bar{S}}) 
	\end{aligned}
\end{equation}

Thus, it follows that the count of assignments for which $\x'_{|S|+1}=0$ that 
satisfy $f'(\x)=1$ precisely equals the number of assignments for which 
$f(\x_S;\x'_{\Bar{S}})=1$. This is, in fact, equivalent to the \emph{local} 
completion count: $C(S,f,\x)$. Put differently, this implies that:

\begin{equation} 
	C(S,f,\x) = \sqrt{\frac{1}{2}\cdot 
		C(\emptyset,f')-2^{2\vert \overline{S}\vert}}
\end{equation}

We now turn our attention to proving the second part of the Lemma. 
Specifically, we show that in the scenario where $f(\x)=0$, the following 
condition is satisfied: 
\begin{equation}
	C(S,f,\x) = \sqrt{\frac{1}{2}\cdot 
		C(\emptyset,f')-2^{2\vert \overline{S}\vert}}
\end{equation}

We will similarly begin by proving that, given $f(\x)=0$, there exist at least 
$2^{|\Bar{S}|}$ vectors $\x'\in\{0,1\}^{|S|+1}$ for which $f'(\x')=0$. 

First, we assume that $\x'_{|\Bar{S}|+1}=1$ (the assignment for feature $|\Bar{S}|+1$ is $1$). Now, for all 
$\x'\in\{0,1\}^{|S|+1}$ such that $\x'_{|\Bar{S}|+1}=1$ it holds that:

\begin{equation}
	\begin{aligned}
		\w'\cdot \x' + b' 
		=\\
		\w'_{\Bar{S}}\cdot \x'_{\Bar{S}}
		+ \w'_{\vert\overline{S}\vert+1} \cdot 
		\x'_{\vert\overline{S}\vert+1}
		+ b'
		=\\
		\w'_{\Bar{S}}\cdot \x'_{\Bar{S}}
		+  \delta'
		+ b'		
		=\\
		\w'_{\Bar{S}}\cdot \x'_{\Bar{S}} 
		+ \left (-(\sum_{i\in \overline{S}}\vert w_{i}\vert) - b' -1\right ) 
		+ b'		
		=\\
		\w'_{\Bar{S}}\cdot \x'_{\Bar{S}} 
		- \sum_{i\in \overline{S}}\vert \w_{i}\vert -1
		< 0
	\end{aligned}
\end{equation}

Given that there are precisely $2^{|\overline{S}|}$ assignments where 
$\x'_{|\overline{S}|+1}=1$, it follows that there exist at least $2^{|\overline{S}|}$ 
assignments for which $f'(\x')=0$. Due to the same reasoning as in the previously discussed case where $f'(\x')=1$, it follows that the subsequent condition is met:

\begin{equation} 
	t =
	2^{\vert \overline{S} \vert} + \sqrt{\frac{1}{2}\cdot 
		C(\emptyset,f')-2^{2\vert \overline{S}\vert}}
\end{equation}

Therefore, the number of assignments, where $\x'_{|S|+1}=1$, that satisfy the 
condition $f'(\x)=0$ is as follows:

\begin{equation} 
	\sqrt{\frac{1}{2}\cdot 
		C(\emptyset,f')-2^{2\vert \overline{S}\vert}}
\end{equation}

Given the aforementioned reasons, we can deduce again that: 
$f'(\x'_{\Bar{S}};\mathbf{0}_{|\Bar{S}|+1})=f(\x_S;\x'_{\Bar{S}})$. 
Consequently, the number of assignments where $\x'_{|S|+1}=0$ and $f'(\x)=0$ 
coincides with those where $f(\x_S;\x'_{\Bar{S}})=0$. This corresponds to the 
\emph{local} completion count: $C(S,f,\x)$ in this context. In other words, it 
again holds that:

\begin{equation} 
	C(S,f,\x) = \sqrt{\frac{1}{2}\cdot 
		C(\emptyset,f')-2^{2\vert \overline{S}\vert}}
\end{equation}

which concludes the reduction.

\begin{lemma}
	G-CC is $\#P$-Complete for MLPs.
\end{lemma}

Proofs of membership and Hardness for Perceptrons will also clearly hold for 
MLPs.

\begin{lemma}
	G-CC is in PTIME for FBDDs.
\end{lemma}

\emph{Proof.} Similarly to the proof of the complexity of \emph{G-CC} for 
Perceptrons (Lemma~\ref{cc_complexity_appendix_perceptrons}), we will assume 
the normalized count $C(S,f)$ which is interchangeable with $c(S,f)$. Each leaf 
node $v$ of $f$ corresponds to some path $\alpha$. We denote by $\alpha_S$ the 
subset of nodes from $\alpha$ that correspond to the features of $S$. We 
suggest the following polynomial algorithm: We enumerate pairs of leaf nodes by iterating over all pairs in its cartesian product. In other words, given that $L:=\{v_1,v_2\}$ includes all the leaf nodes of a function $f$, we iterate over every possible pair of leaf nodes in the set $L:=\{v_1,v_2\}$. This includes the pairs $\{(v_1,v_1), (v_1,v_2), (v_2,v_1), (v_2,v_2)\}$. Given the pair $(v,v')$, 
we perform a counting procedure iff there do not exist two nodes $v_{\alpha}\in 
\alpha_S$ and $v_{\alpha'}\in \alpha'_{S}$ with the same input feature $i$ and 
with \emph{different} path output edges. Intuitively, this means that $\alpha$ and 
$\alpha'$ do not match on the subset $S$. 

We define w.l.o.g. that $v$ corresponds to the counting procedure over 
$\x\in\mathbb{F}$ and that $v'$ corresponds to the counting procedure over 
$\z\in\{0,1\}^{|\overline{S}|}$. Therefore, for each counting procedure, we add 
$2^{n-|\alpha|}\cdot 2^{|\overline{S}|-|\alpha'_{\Bar{S}}|}$. Upon completing 
the iteration across all pairs $(v,v')$, we derive $C(S,f)$. Similarly to our previous assertions, we again note that the expression $2^{n-|\alpha|} \cdot 2^{|\overline{S}|-|\alpha'_{\Bar{S}}|}$ can be computed in polynomial time, since $n$, the number of input features, is given in unary. Furthermore, we observe that the process of iterating through the cartesian product of leaf nodes arises from the definition of the global completion count, which remains unaffected by the order of the features.

\section{Framework Extensions} 
\label{appendix:extension_to_continuous_inputs}

\textbf{Discrete and continuous input and output domains.} To simplify the comprehension of our proofs, we followed common conventions~\cite{BaMoPeSu20, arenas2022computing, waldchen2021computational, ArBaBaPeSu21} and provided them over boolean input and output values. First, we observe that the proofs of duality and uniqueness presented in the initial sections are independent of any assumptions about the input or output domains, making them applicable to both discrete and continuous domains.

Next, we turn our attention to the outcomes of our complexity analysis. It is 
important to note that the analysis we have carried out is not limited to 
binary features but can also be extended to features that take on $k$ possible 
values, where $k$ represents any integer. Moreover, an additional extension can 
adapt our approach to incorporate continuous
inputs. We will now briefly elaborate on the diverse situations in which this 
extension remains relevant.

%many of the results and proofs also hold for 
%the case in which the models handle a \emph{continuous} input space as well.

%\subsection{Multi-Layer Perceptron}
% (such as CSR, MSR, and FR)

Regarding MLP explainability queries, 
earlier research indicates that the complexity of a \emph{satisfiability} 
query on an MLP extends to scenarios involving continuous inputs. Specifically, the work of~\cite{katz2017reluplex} and~\cite{SaLa21} proves that verifying an arbitrary satisfiability query on an MLP with ReLU activations, over a  
continuous input domain, remains NP-complete. The \emph{CSR} query mentioned in this work, when $S:=\emptyset$ is akin to \emph{negating} a satisfiability query, and this implies that the \emph{CSR} query (and consequently also \emph{G-CSR}) in MLPs remains coNP complete for the continuous case as well. We recall that the complexity of the \emph{MSR} and \emph{FR} queries for MLPs are $\Sigma^P_2$-Complete and $\Pi^P_2$-Complete, respectively. This complexity arises from the use of a coNP oracle, which determines whether a subset of features is sufficient, essentially addressing the \emph{CSR} query. Given that \emph{CSR} can also be adapted to handle continuous outputs, the logic applied to \emph{CSR} can similarly be applied to demonstrate that both \emph{MSR} and \emph{FR} queries can be extended to continuous domains.

%In the case of MSR and FR queries, the completeness results hold as well for 
%the continuous case. It is straightforward to derive the proof for 
%\emph{hardness}, 
%as it will be identical to the case of discrete inputs. For proving 
%membership, this will be quite similar as well, only with an oracle that 
%solves in O(1) complexity the co-NP queries for the case of continuous 
%inputs.  
%
%In the case of MLPs, we note that switching from a discrete input space to a 
%continuous one, makes the \emph{local} FN query harder --- and specifically, 
%NP-complete, instead of PTIME for the equivalent query 
%in the case of discrete inputs.

For Perceptrons, the completeness proofs remain valid in a continuous domain for 
\emph{(G)-CSR}, \emph{(G)-MSR}, and \emph{(G)-FR} explainability queries. The continuity of inputs does not 
alter the membership proofs, for the same reasons that hold for MLPs. For hardness proofs, notice that all 
reductions that were derived from the SSP problem, can be adjusted to substitute any call to $\max\{{z_{i},z_{j}}\}$ in our original proof with $\max([z_{i},z_{j}])$.

The \emph{FN} algorithms remain correct for the continuous scenario for Perceptrons and FBDDs, as the insight from theorem~\ref{necessitiy_contrastive_iff} persists. Thus, the algorithm recommended for decision trees remains applicable. For Perceptrons, one can directly determine the minimal and maximal viable values by fixing $S\setminus {i}$ and allowing $i$ to take any possible value. In the context of MLPs, solving the problem remains within polynomial time complexity for discrete inputs, yet transitions to be NP-Complete when dealing with continuous cases. Moreover, it is important to acknowledge that the proofs for the \emph{CC} query also hold only for the discrete case, as its inherent counting nature renders it undefined for the continuous version too.

%\subsection{Perceptron}

%In the case of the global explainability queries with regard to a Perceptron 
%classifier, the completeness proofs still hold for a continuous domain 
%with regard to the CSR, MSR, and FR explainability queries. Input continuity 
%does not change the membership proofs. The hardness proofs can be slightly 
%modified to match this case as well (implying an overall equivalent 
%complexity). This is due to these reductions being done from the classic 
%NP-hard subset-sum problem. A similar reduction holds in the continuous case, 
%as we can equivalently replace all $\max(\{z_{i},z_{j}\})$ appearances with 
%$\max([z_{i},z_{j}])$.
%
%In the case of FN queries for continuous inputs, the same reduction holds as 
%in 
%the 
%general discrete case, and hence the complexity is still PTIME.

%\subsection{FBDD}

%In addition, the proofs for tree classifiers also hold for continuous inputs 
%as 
%well. 

Finally, the proofs that apply to tree classifiers for queries such as \emph{(G)-CSR}, \emph{(G)-MSR}, \emph{(G)-FN}, and \emph{(G)-FR} are equally valid for continuous inputs. This extension to continuous inputs for the local forms mentioned was demonstrated by previous work~\cite{huang2021efficiently}. Given that the global algorithms for decision trees resemble their local counterparts, with the distinction of enumerating pairs of leaf nodes instead of single leaf nodes, the correctness of these algorithms persists in the continuous domain. Consequently, their complexity remains polynomial.

\textbf{Relaxations, probabilistic classification, and regression.} Other possible extensions of our framework might involve alternative, more flexible definitions of our explanation forms, as well as adaptations to different contexts like probabilistic classification or regression.

Possible relaxations of our definitions could incorporate \emph{probabilistic} notions of sufficiency~\cite{waldchen2021computational, izza2021efficient, arenas2022computing, wang2021probabilistic}, restricting them to bounded $\epsilon$-ball domains~\cite{Laagmirhpanikwma21, wu2024verix, izyaxuanjoigalma24, huxuma23}, or to meaningful distributions~\cite{yu2023eliminating, gonirusa22}. Additionally, our definitions could be adapted from the simplified binary classification to regression or probabilistic classification contexts. For instance, in the case of a neural network regression model $f:\mathbb{F}\to \mathbb{R}$, a \emph{sufficient reason} might be defined as a subset $S\subseteq\{1,\ldots,n\}$ of input features such that:

\begin{equation}
    \forall\z\in\mathbb{F} \quad ||f(\x_{S};\z_{\Bar{S}})-f(\x)||_p \leq \delta
\end{equation}

for some $0\leq\delta\leq 1$ and some $\ell_p$-norm. Other concepts explored in our work, such as contrastive reasons, the global versions of our explanation definitions, and the related definitions of necessity and redundancy, can be similarly adapted to this framework.

\section{Terminology and Relationship to Other Explanation Forms}
\label{Terminology_section_appendix}

In this section, we will outline several definitions from the literature that are relevant to those discussed here.

\textbf{Sufficient reasons and abductive explanations.} A sufficient reason is 
also commonly referred to as an \emph{abductive 
explanation}~\cite{ignatiev2019abduction} (abbreviated as \emph{AXP}) or a 
\emph{prime implicant}~\cite{shih2018symbolic} (abbreviated as 
\emph{PI-explanation}). Importantly, our definition of a sufficient reason does 
not automatically imply subset minimality, which is sometimes the case in other 
works. For instance,~\cite{marques2022delivering} describe an AXP as a 
subset-minimal sufficient subset of features, distinguishing it from a 
\emph{weak}-AXP, where subset minimality does not necessarily hold. We follow 
the conventions used by~\cite{BaMoPeSu20}, and define a sufficient reason as 
any sufficient subset of features, and we specify that a subset $S$ is a 
\emph{subset-minimal} sufficient reason only when it is explicitly so.

\textbf{Absloute vs. global sufficient reasons.} Here, we highlight the 
distinctions between \emph{global} sufficient reasons, as discussed throughout 
this work, and the notion of \emph{absolute} sufficient reasons outlined 
in~\cite{ignatiev2019relating}, in which the authors refer to these as 
\emph{absolute explanations}, while~\cite{marques2023logic} define them as 
global abductive explanations. For a given function $f$ and a prediction class 
$c \in \{0,1\}$, an absolute sufficient reason with respect to $\langle f, c 
\rangle$ is a \emph{partial assignment} $\x_S \in \{0,1\}^{|S|}$ to the 
features in $S$, such that:

\begin{equation}
    \forall \z\in\mathbb{F} \ \ [ \ f(\x_{S};\z_{\Bar{S}})=c \ ]
\end{equation}

To determine whether a specific instance qualifies as an \emph{absolute} sufficient reason, we would need to evaluate a \emph{partial assignment} to an input $\x_S$. However, this work centers on comparing local explanations, which apply to a particular instance $\x$, to their corresponding global explanations, which apply to \emph{any} possible instance $\x$. In this context, we concentrate on either the local scenario, which involves subsets of features that are sufficient for determining a specific prediction for $\x$, or the global scenario, where subsets of features are sufficient to determine the prediction for \emph{any} $\x$. As demonstrated by proposition~\ref{necessery_or_redundant_proposition}, this definition of global sufficient reasons also connects to the concepts of necessity and redundancy of features. Specifically, the provably unique subset-minimal global sufficient reason for a function $f$ categorizes features into those that are necessary for a specific $\langle f, \x \rangle$ and those that are globally redundant.

\textbf{Necessity, redundancy, and bias detection.} We follow the notions of feature necessity and redundancy as discussed in~\cite{huang2023feature}, where the focus is primarily on the \emph{local} versions of these explanations. In contrast, our analysis extends to both local variants, which apply to a specific instance $\x$, and global variants, which apply to any instance $\x$. 

These notions of necessity correspond to those discussed 
in~\cite{darwiche2023logic, darwiche2022computation, watson2021local}, while 
the ideas of redundancy are related to fairness and bias, as explored in other 
studies~\cite{darwiche2020reasons, ArBaBaPeSu21, ignatiev2020towards}. There, 
it is often considered that there exists a set $P\subseteq \{1,\ldots,n\}$ of 
protected features that should not influence the prediction. 
Notably,~\cite{ignatiev2020towards} apply the criterion of fairness through 
unawareness (FTU), which involves ensuring that all features in $P$ are 
redundant, whether locally or concerning a specific prediction class. 
Similarly,~\cite{darwiche2020reasons} differentiate between \emph{local} 
biases, termed \emph{prediction} bias, and \emph{global} biases, referred to as 
\emph{classifier} bias. More specifically, a \emph{prediction} $\langle f, \x 
\rangle$ is biased iff:

\begin{equation}
    \exists \z\in\mathbb{F} \ \ [ \ f(\x_{\Bar{P}};\z_{P}) \neq f(\x) \ ]
\end{equation}

which is equivalent to $P$ being \emph{locally} contrastive with respect to $\langle f, \x \rangle$. In the \emph{global} context, a classifier $f$ is deemed biased iff there is at least one input $\x$ where the prediction $\langle f, \x \rangle$ is biased. Equivalently, a classifier $f$ is \emph{unbiased} if and only if all protected features in $P$ are \emph{globally redundant}. A similar notation of bias detection is discussed in~\cite{ArBaBaPeSu21}. 
Moreover,~\cite{audemard2021computational} explore a modified version of this concept, focusing on irrelevancy (redundancy) concerning a particular class.

\textbf{Counting completions and $\delta$-relevant sets.} In this work, we 
investigate the computational complexity of the Count-Completion (\emph{CC}) 
query, a form of explanation also examined in~\cite{BaMoPeSu20}. However, their 
study concentrated on the local variant of this query, whereas we address both 
the local and global variants. 
The \emph{CC} query closely (but not exactly) aligns with a $\delta$-relevant 
set~\cite{waldchen2021computational, izza2021efficient}.
When obtaining a $\delta$-relevant set, the focus is not on calculating the 
completion count itself, but rather on determining whether the completion count 
surpasses a specified threshold $\delta$.

%This can be done by adjustments such as the one presented by Huang et 
%al.~\cite{huang2021efficiently}, proving that in polynomial time it is 
%possible to find all minimal contrastive examples for tree classifiers on 
%continuous, local inputs. In our case, the algorithms 
%are similar to the ones we proposed for the discrete case, with the slight 
%adjustment of covering all \emph{pairs} of leaves --- hence leaving the 
%overall 
%complexity still in PTIME.
%
%which demonstrate how 
%explainability queries of trees can be extended to this 
%scenario.

%\subsection{Counting Completions (CC) in the Continuous Case}
%Finally, we note that there are some cases in which it is not 
%straightforward (and perhaps, interactable) to solve an explainability query 
%for the continuous case. One such example is with the ``counting completions'' 
%(CC) case, in which the current proofs do not hold. Indeed, it is even not 
%straightforward to define \emph{what} is the continuous version of this query. 
%We regard this as interesting future work.

%\begin{proposition}
%G-MCR is \begin{inparaenum}[(i)]
%			\item NP-Complete for MLPs,
%			\item in PTIME for FBDDs and
%   \item in PTIME for Perceptrons
%		\end{inparaenum} 
%\end{proposition}

\end{document}